\documentclass[oupdraft]{bio}

\usepackage{url}
\usepackage{amsmath}
\usepackage{enumerate}
\usepackage{xcolor}
\usepackage{graphicx,psfrag,epsf}
\usepackage{setspace}
\usepackage{parskip}
\usepackage{booktabs}
\usepackage{float}
\usepackage{placeins}
\usepackage{multirow}
\usepackage{makecell}
\usepackage{amssymb}
\usepackage{bbm, dsfont}
\usepackage{subcaption}
\usepackage{caption}
\usepackage{hyperref}
\usepackage{tikz}
\usetikzlibrary{arrows.meta,positioning,fit,calc}
\hypersetup{hidelinks}

\numberwithin{equation}{section}

\makeatletter
\def\ps@plain{%
  \let\@oddfoot\@empty
  \let\@evenfoot\@empty
  \let\@oddhead\@empty
  \let\@evenhead\@empty
}
\makeatother

\begin{document}

\title{A Unified Causal Inference Framework for the Desirability of Outcome Ranking Paradigm in Benefit-Risk Evaluation}

\author{
YUAN FENG$^{1}$, SHIYU SHU$^{1,2}$, YIXIN FANG$^{3}$, IONUT BEBU$^{1,2}$\\
TOSHIMITSU HAMASAKI$^{1,2}$, SCOTT EVANS$^{1,2}$, and GUOQING DIAO$^{1,2,\ast}$\\[4pt]
\textit{$^{1}$Department of Biostatistics and Bioinformatics, The George Washington University, Washington, DC, USA}\\
\textit{$^{2}$The Biostatistics Center, The George Washington University, Washington, DC, USA}\\
\textit{$^{3}$AbbVie Inc., USA}\\[2pt]
{gdiao@email.gwu.edu}
}

\maketitle

\footnotetext{To whom correspondence should be addressed.}

\newpage

\begin{abstract}
{We developed a unified covariate-adjusted causal inference framework for estimating the desirability of outcome ranking (DOOR) probability for benefit-risk evaluation in randomized trials and observational studies. The framework expresses the DOOR probability as a bilinear functional of the marginal ordinal outcome distributions under the two treatment strategies, estimates conditional ordinal distributions through sequential risk-set hazards, and derives the efficient influence function (EIF) of the DOOR probability. The point-estimation simulations compared G-computation, normalized inverse probability weighting (IPW), augmented IPW (AIPW), and targeted maximum likelihood estimation (TMLE), with nuisance functions estimated using generalized linear models or Super Learner (SL). TMLE-SL showed the strongest and most consistent point-estimation performance, with AIPW-SL ranking second. EIF-based inference was then evaluated for AIPW-SL and TMLE-SL, with and without cross-fitting, across settings varying in overlap, treatment-effect heterogeneity, and treatment allocation. CVTMLE-SL showed the strongest overall performance across DOOR-scale bias, recovery of the underlying ordinal distributions, standard-error accuracy, and confidence-interval coverage. We illustrate the methodology using data from the multidrug-resistant organism network of the Antibacterial Resistance Leadership Group.}
{Causal machine learning; Covariate adjustment; Cross-fitting; Desirability of outcome ranking (DOOR); Doubly robust estimation; Efficient influence function; Ordinal outcomes; Targeted maximum likelihood estimation.}
\end{abstract}

\section{Introduction}
\label{sec:intro}

In biomedical research, treatment evaluation often requires balancing efficacy and safety rather than considering each outcome in isolation. Separate analyses of individual efficacy and safety endpoints may fail to capture the overall clinical experience of a patient and can therefore be insufficient for patient-centered benefit-risk assessment \citep{Evans2015DOOR,EvansFollmann2016}. For example, in the phase 3 EPCORE FL-1 trial, adding epcoritamab to lenalidomide and rituximab improved response and progression-free survival in relapsed or refractory follicular lymphoma but also increased high-grade neutropenia and infections, illustrating why efficacy and toxicity should be evaluated jointly rather than through a single endpoint \citep{Falchi2026Epcoritamab}. This challenge has motivated approaches that evaluate treatments through integrated patient-level outcomes rather than collections of marginal endpoint-specific analyses.

The desirability of outcome ranking (DOOR) paradigm was proposed to address this problem by integrating component outcomes into a patient-centric ordinal endpoint based on the desirability of the overall patient response \citep{EvansFollmann2016, Follmann2020OrderedComposite,EvansWuHamasaki2026,HamasakiEvans2025Handbook}. Within this paradigm, prioritized-outcome constructions are specific implementations in which a prespecified hierarchy of component outcomes determines the desirability ranking. In a DOOR analysis, patients are categorized according to an integrated benefit-risk outcome and are compared through the resulting desirability ranking, rather than through isolated endpoint-specific contrasts. This structure is particularly useful in settings where treatment decisions require balancing clinical response, safety events, complications, survival, and quality of life within a single interpretable benefit-risk framework \citep{Hamasaki2025}. 

Once a DOOR endpoint has been prespecified, the DOOR probability provides a population-level treatment comparison with a direct probability-scale interpretation grounded in the patient-centric ranking. It is the probability that a patient under one treatment strategy has a more desirable ranked DOOR outcome than a patient under another strategy, with ties typically assigned weight one half. In randomized trials, this probability can be estimated by comparing treatment groups directly. In observational studies or covariate-adjusted analyses, the same clinical summary can be embedded in a causal framework through the marginal DOOR outcome distributions under the two treatment strategies. 

The next question is how to estimate the DOOR probability efficiently and robustly in practice. Covariate adjustment is relevant in both randomized and observational studies. In randomized trials, adjustment for prognostic baseline covariates can improve precision and power without changing the target estimand \citep{Kahan2014,Diaz2016}. This perspective is reflected in regulatory guidance recommending prespecified covariate adjustment in randomized clinical trials \citep{FDA2023CovAdj,EMA2015CovAdj}. In observational studies, treatment groups may differ systematically in baseline severity and other risk factors, so adjustment is needed to reduce confounding bias under standard causal identification assumptions \citep{RosenbaumRubin1983,Robins1986,RobinsHernanBrumback2000,Fang2024}. G-computation and inverse probability weighting (IPW) rely on the conditional outcome model and treatment model, respectively. Augmented inverse probability weighting (AIPW) and targeted maximum likelihood estimation (TMLE) combine both nuisance components and can be doubly robust and locally efficient under suitable conditions \citep{BangRobins2005,vanDerLaanRubin2006,Hahn1998,ColeHernan2008,Crump2009,Fang2024}. Because doubly robust estimators can still be sensitive when both nuisance working models are mildly misspecified, enhanced doubly robust procedures have also been proposed to improve robustness under model misspecification \citep{KangSchafer2007,YuanYinTan2021}.

For ordinal outcomes, conventional covariate-adjusted analyses often rely on proportional-odds models and odds-ratio-type estimands. However, the proportional-odds assumption is not always assessed, and a common odds ratio may be misleading when treatment effects vary across cumulative cutoffs \citep{Uddin2024,Selman2024}. D{\'i}az, Colantuoni, and Rosenblum addressed this limitation by developing targeted minimum loss-based estimation for ordinal outcomes in randomized trials without imposing a proportional-odds model \citep{Diaz2016}. Nevertheless, the resulting average log-odds ratio remains an ordinal-distribution contrast rather than a directly interpretable pairwise probability, and effects in opposite directions across cutoffs can partially offset one another. Williams, Rosenblum, and D{\'i}az subsequently estimated the Mann--Whitney probability using sequential-hazard TMLE with machine-learning-based outcome regression and cross-fitting in randomized trials \citep{Williams2022Optimising}. When the ordinal categories are placed on a common desirability scale, this probability has the same mathematical form as the DOOR probability. Their work considered this probability in the general setting of a prespecified ordinal outcome. The present work places it within the DOOR paradigm for patient-centric benefit-risk evaluation, in which clinically relevant benefit and risk components define the desirability of the overall patient response, and extends estimation to an unknown, covariate-dependent treatment mechanism.

Viewed more generally, the DOOR probability is a special case of a two-sample causal \(U\)-statistic estimand \citep{Mao2018}. Yin, Yuan, and Tan developed doubly robust estimators for general two-sample causal \(U\)-statistic kernels and emphasized that pairwise causal estimands should be formulated through marginal treatment-specific distributions rather than through the generally unidentified joint distribution of individual-level potential outcomes \citep{YinYuanTan2024}. Their framework includes the DOOR comparison kernel as a special case but models pair-level treatment and outcome mechanisms through monotone single-index models. The finite ordinal setting considered here instead permits a direct factorization through treatment-specific category probabilities and sequential risk-set hazards.

More recently, Shu and colleagues developed IPW, G-computation, and doubly robust estimators for the population-level DOOR probability, with an application to multidrug-resistant organism studies \citep{Shu2026}. That work provides an important foundation for confounding adjustment with DOOR endpoints, but its implementation focused primarily on parametric nuisance working models, including logistic regression for the propensity score and proportional odds regression for the DOOR outcome distribution. Their doubly robust estimator has the AIPW-type form for the marginal category probabilities and does not include the targeted updating step of TMLE. Estimator-specific influence functions for the finite-dimensional nuisance-model parameters were propagated through the marginal category probabilities and then through the DOOR probability by the delta method. This differs from the present approach, which derives the efficient influence function of the DOOR probability itself under the nonparametric observed-data model, rather than relying on a particular finite-dimensional nuisance-model parameterization. 

Reliance on prespecified parametric working models can be sensitive to nonlinear covariate effects, interactions, and violations of proportional-odds-type assumptions. This limitation is practically important in complex biomedical settings, where treatment selection, baseline severity, comorbidities, and outcome distributions may be related in ways that are difficult to prespecify parametrically \citep{Evans2015DOOR,KangSchafer2007}. Flexible nuisance learning provides one way to relax these restrictions, and Super Learner offers a theoretically grounded ensemble approach that combines candidate algorithms in a data-adaptive manner \citep{vanDerLaanPolleyHubbard2007}. Prior simulation work has shown that machine-learning-based singly robust estimators can perform poorly, whereas doubly robust estimators combined with sample splitting and richer learner libraries can improve bias and coverage \citep{Naimi2023MLCausal}. Recent comparative work has also highlighted the practical relevance of combining doubly robust estimation with machine-learning-based nuisance estimation, although the target estimand considered there differs from the DOOR probability studied here \citep{Tan2025}. 

Building on these lines of work, the framework developed here separates the DOOR probability estimand from nuisance-function estimation. Treatment-specific conditional ordinal distributions are represented through sequential risk-set hazards rather than a proportional-odds model \citep{Diaz2016,Williams2022Optimising}, and the resulting marginal distributions are mapped to the DOOR probability through a finite-category bilinear representation. Within this common structure, G-computation and normalized IPW provide singly robust point-estimation benchmarks, whereas AIPW and TMLE provide doubly robust estimation and efficient-influence-function-based inference, including cross-fitted implementations; generalized linear models and Super Learner are used as alternative nuisance-learning procedures. Under an observed-data model that accommodates a covariate-dependent treatment mechanism, we give an explicit finite-category representation of the nonparametric efficient influence function for the DOOR probability. The framework is evaluated in two stages: point-estimation simulations screen all four estimator classes, after which inferential simulations focus on AIPW-SL, TMLE-SL, CVAIPW-SL, and CVTMLE-SL under varying overlap, treatment-effect heterogeneity, and treatment allocation. We illustrate the methodology using data from the Antibacterial Resistance Leadership Group Multi-Drug Resistant Organism network.

The remainder of the article is organized as follows. Section~\ref{sec:meth} defines the DOOR probability estimand and presents the common ordinal-distribution, estimation, and inference framework. Section~\ref{sec:simu} describes the simulation studies. Section~\ref{sec:appl} presents the real-world application. Section~\ref{sec:disc} concludes with a discussion of practical implications, limitations, and future directions.

\section{Methods}
\label{sec:meth}

\subsection{Notation}

Let \(O=(X,A,Y)\) denote a generic observed data unit, where \(X\) is a vector of baseline covariates, \(A\in\{0,1\}\) is a binary treatment indicator, and \(Y\in\{1,\ldots,K\}\) is an ordinal outcome, with larger values representing more desirable outcomes. Independent and identically distributed observations \(O_1,\ldots,O_n\) are sampled from an unknown distribution \(P_0\).

Potential outcomes are denoted by \(Y^1\) and \(Y^0\), representing the ordinal outcomes under treatment and control, respectively. The observed outcome satisfies consistency, \(Y=Y^A\). Define the propensity score
\[
g(a\mid X)=P(A=a\mid X),
\qquad
 a\in\{0,1\},
\]
and the treatment-specific conditional probability mass function
\[
Q_a(k\mid X)
=
P(Y=k\mid A=a,X),
\qquad
 a\in\{0,1\},\quad k=1,\ldots,K.
\]

\subsection{DOOR probability estimand and identification}
\label{subsec:door_estimand}

The causal estimand of interest is a population-level pairwise comparison of the ranked outcomes under two treatment strategies. Let \(Y^{1,\dagger}\) denote a random draw from the marginal distribution of \(Y^1\), and let \(Y^{0,\ddagger}\) denote an independent random draw from the marginal distribution of \(Y^0\). The causal DOOR probability is
\begin{equation}
\psi
=
P(Y^{1,\dagger}>Y^{0,\ddagger})
+
\tfrac12 P(Y^{1,\dagger}=Y^{0,\ddagger}),
\label{eq:door_def}
\end{equation}
with ties receiving weight one half. This formulation is a Mann--Whitney-type causal estimand based on two independently selected members of the target population \citep{Zhang2019MW}.

A same-individual contrast would instead compare \(Y^1\) and \(Y^0\) for the same person. Such a contrast generally depends on the joint distribution of \((Y^1,Y^0)\), which is not identified from the observed data without additional, untestable restrictions on their dependence \citep{FayLi2024}. In contrast, \eqref{eq:door_def} depends only on the two marginal potential-outcome distributions \citep{Zhang2019MW,YinYuanTan2024}. 

Let
\[
p_k^a
=
P(Y^a=k),
\qquad
 a\in\{0,1\},\quad k=1,\ldots,K,
\]
and define
\[
p^a=(p_1^a,\ldots,p_K^a)^\top.
\]
Expanding \eqref{eq:door_def} over the \(K^2\) possible category pairs gives
\begin{equation}
\psi
=
\sum_{k=1}^K
\sum_{\ell=1}^K
\left\{
\mathbbm{1}(k>\ell)
+
\tfrac12\mathbbm{1}(k=\ell)
\right\}
p_k^1p_\ell^0.
\label{eq:door_pair_sum}
\end{equation}
Define the \(K\times K\) comparison matrix
\[
M_{k\ell}
=
\mathbbm{1}(k>\ell)
+
\tfrac12\mathbbm{1}(k=\ell).
\]
Then
\begin{equation}
\psi
=
p^{1\top}M p^0.
\label{eq:door_bilinear}
\end{equation}

Identification therefore reduces to identifying \(p^1\) and \(p^0\), rather than the joint distribution of \((Y^1,Y^0)\). An equivalent bilinear matrix representation was used by \citet{Shu2026}. We invoke four standard assumptions.

\noindent\textbf{(A1) Consistency.} The observed outcome satisfies \(Y=Y^A\).

\noindent\textbf{(A2) No interference.} An individual's potential outcome under a given treatment does not depend on the treatment assignments of other individuals.

\noindent\textbf{(A3) Conditional exchangeability.}
\[
(Y^1,Y^0)\perp\!\!\!\perp A\mid X.
\]

\noindent\textbf{(A4) Positivity.}
\[
0<P(A=1\mid X)<1
\quad\text{almost surely}.
\]

Conditional exchangeability holds by design in randomized trials and is assumed after adjustment for measured baseline covariates in observational studies. Under assumptions (A1)--(A4),

\begin{equation}
p_k^a
=
\mathbb{E}\{Q_a(k\mid X)\},
\qquad
 a\in\{0,1\},\quad k=1,\ldots,K,
\label{eq:pk_ident}
\end{equation}
where the expectation is taken with respect to the marginal distribution of \(X\) in the target population.

\subsection{Treatment-specific ordinal distribution framework}
\label{subsec:ordinal_estimation}

The estimation procedures in this framework share a common construction. Each procedure first estimates the treatment-specific marginal cumulative probabilities
\[
\theta_k^a
=
P(Y^a\le k),
\qquad
a\in\{0,1\},\quad k=1,\ldots,K-1,
\]
and then converts these estimates to treatment-specific marginal category probabilities before applying the DOOR map. We use \(m\) to index the estimation procedure. The procedures considered are G-computation, denoted by GCOMP in subscripts, IPW, AIPW, and TMLE, together with the cross-fitted AIPW and TMLE procedures introduced in Section~\ref{subsec:crossfit}. For later use, define
\[
B_{ik}
=
\mathbbm{1}\{Y_i\le k\}.
\]

For procedures that use an outcome regression, the treatment-specific conditional ordinal distribution is represented through the sequential risk-set hazards of \citet{Diaz2016}. Define
\[
R_{ik}
=
\mathbbm{1}\{Y_i\ge k\},
\qquad
Z_{ik}
=
\mathbbm{1}\{Y_i=k\},
\]
and
\[
h_{a,k}(X)
=
P(Z_{ik}=1\mid R_{ik}=1,A_i=a,X_i=X).
\]
The hazards induce the conditional cumulative probabilities
\begin{equation}
F_{a,k}(X)
=
P(Y\le k\mid A=a,X)
=
1-
\prod_{j=1}^k
\{1-h_{a,j}(X)\}.
\label{eq:theta_from_hazard}
\end{equation}
The conditional category probabilities are recovered as
\[
\begin{aligned}
Q_a(1\mid X)
&=
F_{a,1}(X),\\
Q_a(k\mid X)
&=
F_{a,k}(X)-F_{a,k-1}(X),
\qquad k=2,\ldots,K-1,\\
Q_a(K\mid X)
&=
1-F_{a,K-1}(X).
\end{aligned}
\]
Consequently,
\[
\theta_k^a
=
\mathbb{E}\{F_{a,k}(X)\},
\qquad
p_k^a
=
\mathbb{E}\{Q_a(k\mid X)\}.
\]
The sequential hazard representation yields nonnegative conditional category probabilities that sum to one whenever the fitted hazards lie in \([0,1]\).

Let \(\hat\theta_{k,m}^a\) denote the estimate of \(\theta_k^a\) produced by procedure \(m\). The corresponding marginal category-probability estimates are
\begin{equation}
\hat p_{k,m}^a
=
\begin{cases}
\hat\theta_{1,m}^a,
& k=1,\\[3pt]
\hat\theta_{k,m}^a-\hat\theta_{k-1,m}^a,
& k=2,\ldots,K-1,\\[3pt]
1-\hat\theta_{K-1,m}^a,
& k=K.
\end{cases}
\label{eq:theta_to_p}
\end{equation}
Define
\[
\hat p_m^a
=
(\hat p_{1,m}^a,\ldots,\hat p_{K,m}^a)^\top,
\]
and let \(\hat\theta_m\) collect the \(2(K-1)\) treatment-specific cumulative-probability estimates. The common final map is
\begin{equation}
\hat\theta_m
\longmapsto
(\hat p_m^1,\hat p_m^0)
\longmapsto
\hat\psi_m
=
\hat p_m^{1\top}M\hat p_m^0.
\label{eq:common_door_map}
\end{equation}

The treatment mechanism and ordinal hazards are estimated using either generalized linear models or Super Learner. Fitted propensity scores are truncated to \([0.001,0.999]\) for all procedures. Whenever fitted ordinal hazards are used, they are truncated to the same range, with the bounds reapplied after each TMLE targeting update.

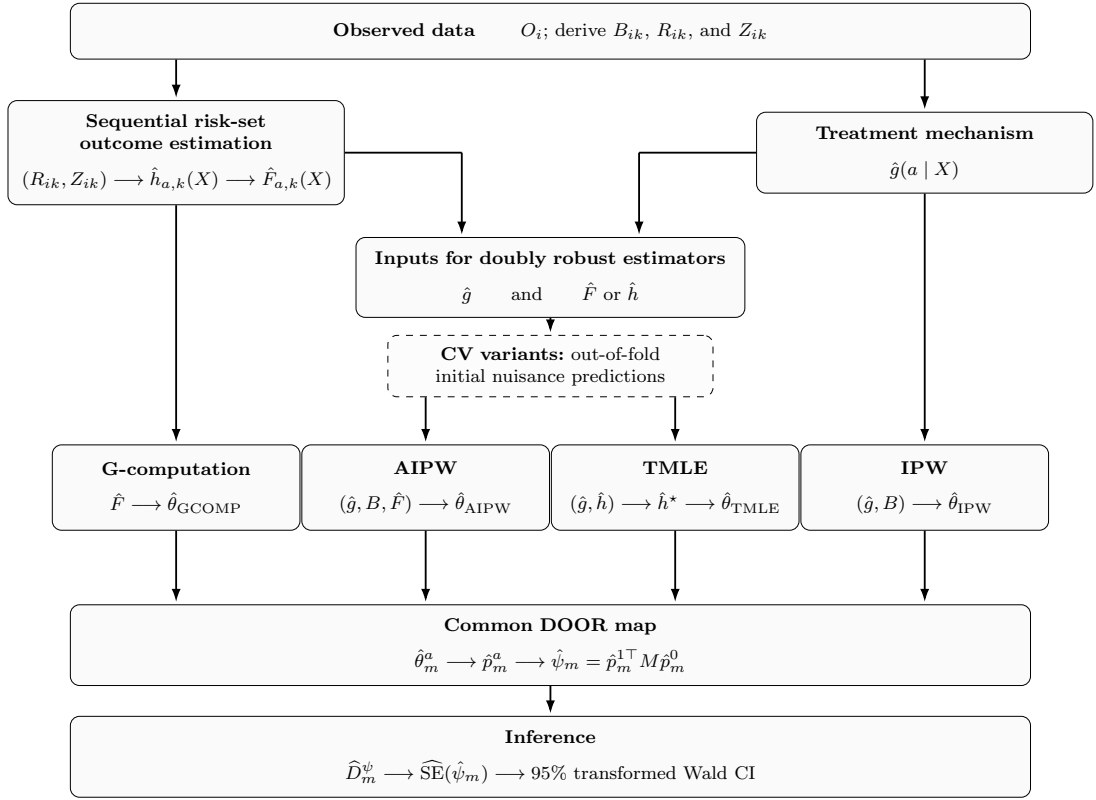
\begin{figure}[!tbp]
\centering
\makebox[\linewidth][c]{%
\resizebox{0.98\linewidth}{!}{%
\begin{tikzpicture}[
  font=\footnotesize,
  topbox/.style={
    draw,
    rounded corners,
    align=center,
    inner sep=6pt,
    text width=14.2cm,
    minimum height=0.85cm,
    fill=gray!4
  },
  inputbox/.style={
    draw,
    rounded corners,
    align=center,
    inner sep=6pt,
    text width=4.7cm,
    minimum height=1.10cm,
    fill=gray!4
  },
  jointbox/.style={
    draw,
    rounded corners,
    align=center,
    inner sep=6pt,
    text width=5.5cm,
    minimum height=0.85cm,
    fill=gray!4
  },
  methodbox/.style={
    draw,
    rounded corners,
    align=center,
    inner sep=6pt,
    text width=3.35cm,
    minimum height=1.30cm,
    fill=gray!4
  },
  commonbox/.style={
    draw,
    rounded corners,
    align=center,
    inner sep=6pt,
    text width=14.2cm,
    minimum height=0.85cm,
    fill=gray!4
  },
  cvbox/.style={
    draw,
    dashed,
    rounded corners,
    align=center,
    inner sep=5pt,
    text width=4.6cm,
    minimum height=0.72cm,
    fill=gray!2
  },
  plainline/.style={
    thick
  },
  downarrow/.style={
    -{Latex[length=2mm,width=1.4mm]},
    thick,
    shorten >=1pt
  }
]

\node[topbox] (data) at (0,0) {
\textbf{Observed data}
\qquad
\(O_i\);
derive \(B_{ik}\), \(R_{ik}\), and \(Z_{ik}\)
};

\node[inputbox] (outcome) at (-5.7,-1.85) {
\textbf{Sequential risk-set\\ outcome estimation}
\[
(R_{ik},Z_{ik})
\longrightarrow
\hat h_{a,k}(X)
\longrightarrow
\hat F_{a,k}(X)
\]
};

\node[inputbox] (ps) at (5.7,-1.85) {
\textbf{Treatment mechanism}
\[
\hat g(a\mid X)
\]
};

\node[jointbox] (joint) at (0,-3.75) {
\textbf{Inputs for doubly robust estimators}
\[
\hat g
\qquad\text{and}\qquad
\hat F\ \text{or}\ \hat h
\]
};

\node[cvbox] (cv) at (0,-5.10) {
\textbf{CV variants:}
out-of-fold initial nuisance predictions
};

\node[methodbox] (gcomp) at (-5.7,-6.95) {
\textbf{G-computation}
\[
\hat F
\longrightarrow
\hat\theta_{\mathrm{GCOMP}}
\]
};

\node[methodbox] (aipw) at (-1.9,-6.95) {
\textbf{AIPW}
\[
(\hat g,B,\hat F)
\longrightarrow
\hat\theta_{\mathrm{AIPW}}
\]
};

\node[methodbox] (tmle) at (1.9,-6.95) {
\textbf{TMLE}
\[
(\hat g,\hat h)
\longrightarrow
\hat h^\star
\longrightarrow
\hat\theta_{\mathrm{TMLE}}
\]
};

\node[methodbox] (ipw) at (5.7,-6.95) {
\textbf{IPW}
\[
(\hat g,B)
\longrightarrow
\hat\theta_{\mathrm{IPW}}
\]
};

\node[commonbox] (map) at (0,-9.35) {
\textbf{Common DOOR map}
\[
\hat\theta_m^a
\longrightarrow
\hat p_m^a
\longrightarrow
\hat\psi_m
=
\hat p_m^{1\top}M\hat p_m^0
\]
};

\node[commonbox] (infer) at (0,-11.05) {
\textbf{Inference}
\[
\widehat D_m^\psi
\longrightarrow
\widehat{\operatorname{SE}}(\hat\psi_m)
\longrightarrow
95\%\ \text{transformed Wald CI}
\]
};

\draw[downarrow]
([xshift=-5.7cm]data.south)
--
(outcome.north);

\draw[downarrow]
([xshift=5.7cm]data.south)
--
(ps.north);

\draw[downarrow]
(outcome.south)
--
(gcomp.north);

\draw[downarrow]
(ps.south)
--
(ipw.north);

\coordinate (outsource) at (outcome.east);
\coordinate (pssource) at (ps.west);

\coordinate (jointleft) at ([xshift=-1.35cm]joint.north);
\coordinate (jointright) at ([xshift=1.35cm]joint.north);

\coordinate (outcorner) at (jointleft |- outsource);
\coordinate (pscorner) at (jointright |- pssource);

\draw[plainline]
(outsource)
--
(outcorner);

\draw[downarrow]
(outcorner)
--
(jointleft);

\draw[plainline]
(pssource)
--
(pscorner);

\draw[downarrow]
(pscorner)
--
(jointright);

\draw[downarrow]
(joint.south)
--
(cv.north);

\coordinate (cvtoaipw) at (aipw.north |- cv.south);
\coordinate (cvtotmle) at (tmle.north |- cv.south);

\draw[downarrow]
(cvtoaipw)
--
(aipw.north);

\draw[downarrow]
(cvtotmle)
--
(tmle.north);

\coordinate (mapgcomp) at (gcomp.south |- map.north);
\coordinate (mapaipw) at (aipw.south |- map.north);
\coordinate (maptmle) at (tmle.south |- map.north);
\coordinate (mapipw) at (ipw.south |- map.north);

\draw[downarrow]
(gcomp.south)
--
(mapgcomp);

\draw[downarrow]
(aipw.south)
--
(mapaipw);

\draw[downarrow]
(tmle.south)
--
(maptmle);

\draw[downarrow]
(ipw.south)
--
(mapipw);

\draw[downarrow]
(map.south)
--
(infer.north);

\end{tikzpicture}
}}
\caption{Workflow for covariate-adjusted estimation of the DOOR probability. G-computation and IPW are singly robust point estimators based on the outcome regression and treatment mechanism, respectively. AIPW and TMLE use both nuisance functions. CVAIPW and CVTMLE use out-of-fold initial nuisance predictions, with CVTMLE subsequently applying the pooled targeting step. All estimators use the same map from the marginal cumulative probabilities under the two treatment strategies to the DOOR probability. Symbols are defined in the main text.}
\label{fig:method_workflow}
\par\smallskip
\begin{minipage}{0.92\linewidth}
\footnotesize
\textit{Alt text:} Flow diagram showing how the treatment and ordinal-outcome models feed the six estimators, which estimate treatment-specific marginal distributions before these are mapped to the DOOR probability.
\end{minipage}
\par
\end{figure}

\subsection{G-computation}
\label{subsec:gcomp}

G-computation estimates the treatment-specific marginal cumulative
probabilities by averaging the fitted conditional cumulative probabilities
over the empirical covariate distribution:
\begin{equation}
\hat\theta_{k,\mathrm{GCOMP}}^a
=
\frac{1}{n}
\sum_{i=1}^n
\hat F_{a,k}(X_i).
\label{eq:theta_gcomp}
\end{equation}
The estimated category probabilities and DOOR probability are then obtained
from \eqref{eq:theta_to_p} and \eqref{eq:common_door_map}. This is a singly
robust estimator whose consistency depends on consistent estimation of the
conditional ordinal outcome distribution. The construction follows the g-formula principle of \citet{Robins1986}; related parametric G-computation estimators for the DOOR probability were considered by \citet{Shu2026}, and a general treatment of outcome-regression-based causal estimation is provided by \citet{Fang2024}.

\subsection{Inverse probability weighting}
\label{subsec:ipw}

For each treatment arm \(a\) and cutoff \(k\), we use the normalized
H\'ajek-type IPW estimator
\begin{equation}
\hat\theta_{k,\mathrm{IPW}}^a
=
\frac{
\sum_{i=1}^n
\dfrac{\mathbbm{1}\{A_i=a\}}
{\hat g(a\mid X_i)}
B_{ik}
}{
\sum_{i=1}^n
\dfrac{\mathbbm{1}\{A_i=a\}}
{\hat g(a\mid X_i)}
}.
\label{eq:theta_ipw}
\end{equation}
The estimated category probabilities and DOOR probability are obtained
from \eqref{eq:theta_to_p} and \eqref{eq:common_door_map}. This is a singly
robust estimator whose consistency depends on consistent estimation of the
treatment mechanism. The H\'ajek normalization produces a weighted mean on
the probability scale and can improve finite-sample stability relative to
the unnormalized Horvitz--Thompson estimator
\citep{RobinsHernanBrumback2000,LuncefordDavidian2004,ColeHernan2008}.
The OptPS estimators use the same formula with the alternative
outcome-oriented propensity-score covariate specification.

\subsection{Efficient influence function for the DOOR probability}
\label{subsec:if_framework}

Influence-function-based inference is developed for AIPW, TMLE, and their
cross-fitted variants. For each treatment arm \(a\) and cutoff \(k\), the
efficient influence function for the marginal cumulative probability
\(\theta_k^a=P(Y^a\le k)\) under the nonparametric observed-data model is
\begin{equation}
D_{\theta,k}^{a,\mathrm{EIF}}(O)
=
\frac{\mathbbm{1}\{A=a\}}
{g(a\mid X)}
\left\{
B_k-F_{a,k}(X)
\right\}
+
F_{a,k}(X)-\theta_k^a,
\label{eq:eif_theta}
\end{equation}
where \(B_k=\mathbbm{1}\{Y\le k\}\)
\citep{Kennedy2022DRDML,Fang2024}. The corresponding efficient influence
function for the marginal category probability \(p_k^a=P(Y^a=k)\) is
\begin{equation}
D_{p,k}^{a,\mathrm{EIF}}(O)
=
\frac{\mathbbm{1}\{A=a\}}
{g(a\mid X)}
\left[
\mathbbm{1}\{Y=k\}
-
Q_a(k\mid X)
\right]
+
Q_a(k\mid X)-p_k^a.
\label{eq:eif_pk}
\end{equation}
Equivalently, the category-probability influence functions can be obtained
from the cumulative-probability influence functions by adjacent
differencing.

The DOOR probability is the bilinear map
\[
\psi
=
\Phi(p^1,p^0)
=
p^{1\top}Mp^0.
\]
For perturbations \(\delta^1\) and \(\delta^0\), its derivative is
\begin{equation}
\left.
\frac{d}{dt}
\Phi(p^1+t\delta^1,p^0+t\delta^0)
\right|_{t=0}
=
(\delta^1)^\top Mp^0
+
p^{1\top}M\delta^0.
\label{eq:phi_derivative}
\end{equation}
Here \(p^1\), \(p^0\), and \(\psi\) are functionals of the observed-data
distribution. Because the component functionals are pathwise
differentiable and \(\Phi\) is continuously differentiable, the functional
delta method applies the derivative in \eqref{eq:phi_derivative} to their
efficient influence functions \citep{vanDerVaart1998}.

Let
\[
D_p^{a,\mathrm{EIF}}(O)
=
\left\{
D_{p,1}^{a,\mathrm{EIF}}(O),
\ldots,
D_{p,K}^{a,\mathrm{EIF}}(O)
\right\}^{\top}.
\]
The efficient influence function for the DOOR probability is therefore
\begin{equation}
\begin{aligned}
D^{\psi,\mathrm{EIF}}(O)
&=
D_p^{1,\mathrm{EIF}}(O)^\top Mp^0
+
p^{1\top}M D_p^{0,\mathrm{EIF}}(O)\\
&=
\sum_{k=1}^K
\sum_{\ell=1}^K
M_{k\ell}
\left\{
p_\ell^0D_{p,k}^{1,\mathrm{EIF}}(O)
+
p_k^1D_{p,\ell}^{0,\mathrm{EIF}}(O)
\right\}.
\end{aligned}
\label{eq:eif_psi}
\end{equation}
Equation~\eqref{eq:eif_psi} is the influence-function representation used
for the doubly robust estimators below.

\subsection{Augmented inverse probability weighting}
\label{subsec:aipw}

AIPW augments the outcome-regression plug-in estimator with the residual
term from the efficient influence function. For each treatment arm and
cutoff,
\begin{equation}
\hat\theta_{k,\mathrm{AIPW}}^a
=
\frac{1}{n}
\sum_{i=1}^n
\left[
\frac{\mathbbm{1}\{A_i=a\}}
{\hat g(a\mid X_i)}
\left\{
B_{ik}-\hat F_{a,k}(X_i)
\right\}
+
\hat F_{a,k}(X_i)
\right].
\label{eq:theta_aipw}
\end{equation}
Equivalently, AIPW is the one-step estimator obtained by adding the
empirical efficient-influence-function correction to the
G-computation plug-in estimator. Its estimated cumulative-probability
influence-function contribution is
\begin{equation}
\widehat D_{\theta,k}^{a,\mathrm{AIPW}}(O_i)
=
\frac{\mathbbm{1}\{A_i=a\}}
{\hat g(a\mid X_i)}
\left\{
B_{ik}-\hat F_{a,k}(X_i)
\right\}
+
\hat F_{a,k}(X_i)
-
\hat\theta_{k,\mathrm{AIPW}}^a.
\label{eq:if_theta_aipw}
\end{equation}

Under standard regularity conditions, AIPW is consistent if either the
propensity score or the conditional cumulative outcome regression is
consistently estimated and is locally efficient when both are consistently
estimated \citep{BangRobins2005,Hahn1998,Fang2024}. The AIPW estimator of
the DOOR probability is
\[
\hat\psi_{\mathrm{AIPW}}
=
\hat p_{\mathrm{AIPW}}^{1\top}
M
\hat p_{\mathrm{AIPW}}^0.
\]
AIPW is not a substitution estimator and therefore does not automatically
enforce monotonicity of the estimated marginal cumulative probabilities in
finite samples.

\subsection{Targeted maximum likelihood estimation}
\label{subsec:tmle}

TMLE updates the initial ordinal hazard estimates through a targeted
substitution procedure \citep{vanDerLaanRubin2006,vanDerLaanRose2011}.
Following \citet{Diaz2016}, targeting is performed on the hazard scale for
the vector of treatment-specific cumulative probabilities.

Let \(\hat h_{a,j}^{(t)}(X)\) and \(\hat F_{a,k}^{(t)}(X)\) denote the
hazard and cumulative-probability estimates at iteration \(t\). For
\(a\in\{0,1\}\) and \(j,k=1,\ldots,K-1\), define the clever covariates
\begin{equation}
H_{ak}^{(t)}(j,X,A)
=
\frac{\mathbbm{1}\{A=a\}}
{\hat g(a\mid X)}
\frac{
1-\hat F_{a,k}^{(t)}(X)
}{
1-\hat F_{a,j}^{(t)}(X)
}
\mathbbm{1}\{j\le k\}.
\label{eq:clever_covariate}
\end{equation}
The long-format logistic fluctuation model is
\begin{equation}
\operatorname{logit}
\{\hat h_j^{(t)}(\epsilon)(X,A)\}
=
\operatorname{logit}
\{\hat h_j^{(t)}(X,A)\}
+
\sum_{a=0}^1
\sum_{k=1}^{K-1}
\epsilon_{ak}
H_{ak}^{(t)}(j,X,A),
\label{eq:tmle_fluctuation}
\end{equation}
where
\[
\hat h_j^{(t)}(X,A)
=
\sum_{a=0}^1
\mathbbm{1}\{A=a\}
\hat h_{a,j}^{(t)}(X).
\]
The fluctuation parameter is estimated among observations in the relevant
risk sets, using the current logit hazard as an offset. The fitted
fluctuation defines the updated hazards
\[
\hat h_{a,j}^{(t+1)}(X)
=
\hat h_j^{(t)}(\hat\epsilon)(X,a).
\]
The updated hazards are used to recompute the cumulative probabilities and
the clever covariates, and the procedure is iterated until the empirical
mean squared change in the hazard predictions is no greater than
\(10^{-4}/n\), or until 100 iterations are reached.

After convergence, the targeted hazards
\(\hat h_{a,j}^{\star}(X)\) induce
\(\hat F_{a,k}^{\star}(X)\) through
\eqref{eq:theta_from_hazard}. The targeted marginal cumulative
probabilities are
\begin{equation}
\hat\theta_{k,\mathrm{TMLE}}^a
=
\frac{1}{n}
\sum_{i=1}^n
\hat F_{a,k}^{\star}(X_i),
\label{eq:theta_tmle}
\end{equation}
and the estimated cumulative-probability influence-function contribution is
\begin{equation}
\widehat D_{\theta,k}^{a,\mathrm{TMLE}}(O_i)
=
\frac{\mathbbm{1}\{A_i=a\}}
{\hat g(a\mid X_i)}
\left\{
B_{ik}-\hat F_{a,k}^{\star}(X_i)
\right\}
+
\hat F_{a,k}^{\star}(X_i)
-
\hat\theta_{k,\mathrm{TMLE}}^a.
\label{eq:if_theta_tmle}
\end{equation}
The fluctuation score equations make the empirical means of these
influence-function contributions approximately zero. Finally,
\[
\hat\psi_{\mathrm{TMLE}}
=
\hat p_{\mathrm{TMLE}}^{1\top}
M
\hat p_{\mathrm{TMLE}}^0.
\]
Because TMLE is a substitution estimator based on the targeted hazards, it
preserves a coherent treatment-specific ordinal distribution before the
DOOR map is applied.

\subsection{Cross-fitted AIPW and TMLE}
\label{subsec:crossfit}

We also consider cross-fitted versions of the two doubly robust estimators.
The sample is partitioned into \(V\) mutually exclusive folds. For each
validation fold, the propensity score and all treatment-specific ordinal
hazards are fitted using observations outside that fold and are then
predicted for observations in the held-out fold. Stacking these predictions
produces out-of-fold estimates
\(\hat g^{\mathrm{cf}}\), \(\hat h^{\mathrm{cf}}\), and
\(\hat F^{\mathrm{cf}}\) for every observation.

For point estimation, CVAIPW substitutes the stacked out-of-fold propensity-score and cumulative-probability predictions into \eqref{eq:theta_aipw}, whereas CVTMLE uses the stacked out-of-fold propensity-score and ordinal-hazard predictions as initial estimates and applies a single hazard-scale targeting update to the stacked long-format data using \eqref{eq:clever_covariate} and \eqref{eq:tmle_fluctuation}. The updated hazards determine the treatment-specific cumulative and category probabilities before the DOOR map is applied. For inference, the influence-function contributions are evaluated using the out-of-fold nuisance estimates for CVAIPW and the out-of-fold propensity scores together with the updated hazards for CVTMLE. This construction reduces reliance on restrictive empirical-process conditions when flexible nuisance learners are used \citep{Kennedy2022DRDML,ZhengVanDerLaan2010CVTMLE,Smith2025CVTMLE}.

\subsection{Standard errors and confidence intervals}
\label{subsec:inference}

Influence-function-based inference is implemented for
\[
m
\in
\{
\mathrm{AIPW},
\mathrm{TMLE},
\mathrm{CVAIPW},
\mathrm{CVTMLE}
\}.
\]
Let
\[
\widehat D_{\theta,m}^a(O_i)
=
\left\{
\widehat D_{\theta,1,m}^a(O_i),
\ldots,
\widehat D_{\theta,K-1,m}^a(O_i)
\right\}^{\top}
\]
denote the estimated cumulative-probability influence-function vector for
procedure \(m\). The corresponding category-probability contributions are
obtained by adjacent differencing:
\[
\begin{aligned}
\widehat D_{p,1,m}^a(O_i)
&=
\widehat D_{\theta,1,m}^a(O_i),\\
\widehat D_{p,k,m}^a(O_i)
&=
\widehat D_{\theta,k,m}^a(O_i)
-
\widehat D_{\theta,k-1,m}^a(O_i),
\qquad k=2,\ldots,K-1,\\
\widehat D_{p,K,m}^a(O_i)
&=
-\widehat D_{\theta,K-1,m}^a(O_i).
\end{aligned}
\]
Let
\[
\widehat D_{p,m}^a(O_i)
=
\left\{
\widehat D_{p,1,m}^a(O_i),
\ldots,
\widehat D_{p,K,m}^a(O_i)
\right\}^{\top}.
\]
Applying the functional delta-method map in
\eqref{eq:eif_psi}, the estimated influence-function contribution for the
DOOR probability is
\begin{equation}
\widehat D_m^\psi(O_i)
=
\widehat D_{p,m}^1(O_i)^\top
M\hat p_m^0
+
\hat p_m^{1\top}
M\widehat D_{p,m}^0(O_i).
\label{eq:door_if}
\end{equation}

Define
\[
\overline D_m^\psi
=
\frac{1}{n}
\sum_{i=1}^n
\widehat D_m^\psi(O_i).
\]
The estimated standard error is
\begin{equation}
\widehat{\operatorname{SE}}_{\psi,m}
=
\left\{
\frac{1}{n(n-1)}
\sum_{i=1}^n
\left[
\widehat D_m^\psi(O_i)
-
\overline D_m^\psi
\right]^2
\right\}^{1/2}.
\label{eq:se_psi}
\end{equation}

Because the DOOR probability is bounded between zero and one, confidence
intervals are constructed on the inverse-hyperbolic-tangent scale,
\begin{equation}
z
=
\operatorname{atanh}(2\psi-1)
=
\frac12
\log
\left\{
\frac{\psi}{1-\psi}
\right\}.
\label{eq:atanh_transform}
\end{equation}
Related transformed intervals have been considered for
Wilcoxon--Mann--Whitney-type pairwise probability measures
\citep{SimonoffHochbergReiser1988,Edwardes1995}. By the delta method
\citep{vanDerVaart1998},
\[
\frac{d}{d\psi}
\operatorname{atanh}(2\psi-1)
=
\frac{1}{2\psi(1-\psi)},
\]
and hence
\begin{equation}
\widehat{\operatorname{SE}}_{z,m}
=
\frac{
\widehat{\operatorname{SE}}_{\psi,m}
}{
2\hat\psi_m(1-\hat\psi_m)
}.
\label{eq:atanh_se}
\end{equation}
A nominal \(95\%\) confidence interval is formed as
\[
\hat z_m
\pm
z_{0.975}
\widehat{\operatorname{SE}}_{z,m},
\qquad
\hat z_m
=
\operatorname{atanh}(2\hat\psi_m-1),
\]
and is mapped back to the probability scale using
\[
\psi
=
\frac{\exp(2z)}{1+\exp(2z)}
=
\frac{1+\tanh(z)}{2}.
\]
All reported influence-function-based confidence intervals use this
transformed Wald construction.

\section{Simulations}
\label{sec:simu}

The simulation study had two components. The point-estimation component
compared G-computation, IPW, AIPW, and TMLE under the original
treatment-assignment mechanisms at \(n=2000\), with \(n=500\) used as a
finite-sample sensitivity analysis. These analyses varied the covariate
representation, propensity-score overlap, treatment-effect structure, and
nuisance-learning strategy.

The inferential component focused on AIPW-SL, TMLE-SL, CVAIPW-SL, and
CVTMLE-SL in the X-working setting with \(n=2000\). It jointly varied
propensity-score overlap, treatment-effect heterogeneity, and treatment-
allocation balance. The ordinal endpoint had \(K=4\) ordered categories,
where 1 denotes the least desirable and 4 the most desirable outcome.
Unless otherwise stated, each scenario used \(R=1000\) Monte Carlo
replications.

\subsection{Data-generating mechanism}
\label{subsec:sim_dgm}

The data-generating mechanism was adapted from \citet{Tan2025}. Baseline covariates were generated as
\[
X=(X_1,\ldots,X_9)^\top\sim N(0,I_9),
\]
and nonlinear transformed covariates \(W=\phi(X)\) were defined by
\[
\begin{aligned}
W
=\Bigl(&
\exp(X_1/2),
\exp(X_2/3),
X_3^2,
X_4^2,
X_5,
X_6,\\
&X_7+X_8,
X_7^2+X_8^2,
X_9^3
\Bigr)^\top,
\end{aligned}
\]
followed by componentwise standardization.

Treatment assignment was generated as
\[
A\sim\operatorname{Bernoulli}\{e(W)\},
\qquad
e(W)=\operatorname{expit}\{\eta(W)\},
\]
where
\[
\eta(W)
=
\begin{cases}
(-3-W_1+2W_2-3W_3+3W_4+2W_5+W_6)/15,
&\text{large overlap},\\[4pt]
(-8W_1+1.5W_2+0.5W_3-0.5W_4+2.5W_5-0.5W_6)/5,
&\text{small overlap}.
\end{cases}
\]
The small-overlap mechanism produces more extreme treatment probabilities and was included to evaluate sensitivity to limited overlap and variable inverse-probability weights.

\begin{figure}[!tbp]
\centering
\makebox[\linewidth][c]{%
\includegraphics[width=0.94\linewidth]{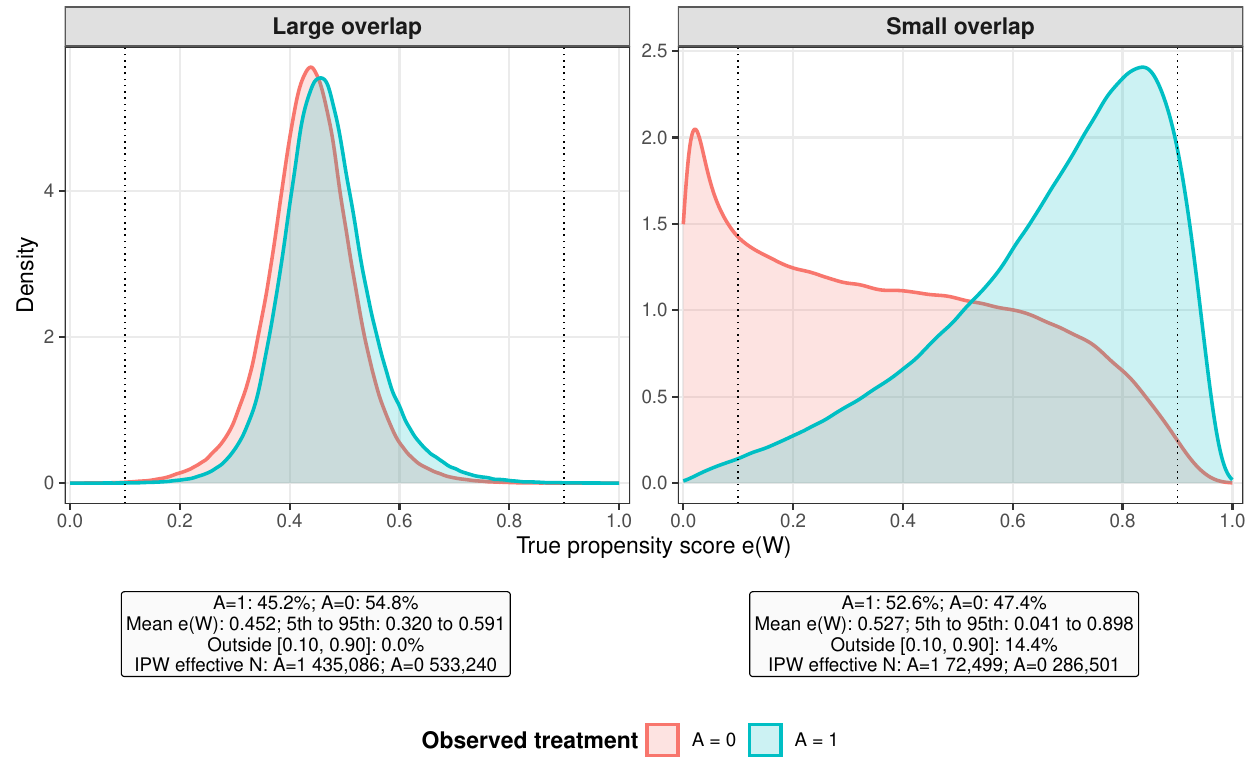}
}
\caption{True propensity-score distributions and treatment-assignment summaries under the large- and small-overlap simulation regimes. Each panel is based on a separate Monte Carlo population of size \(10^6\). The summary strip reports the induced treatment allocation, selected propensity-score distribution summaries, the proportion of individuals with propensity scores outside \([0.10,0.90]\), and the inverse-probability-weighted effective sample size within each treatment group.}
\label{fig:sim_overlap}
\par\smallskip
\begin{minipage}{0.92\linewidth}
\footnotesize
\textit{Alt text:} Two panels compare the true propensity-score distributions under the large- and small-overlap regimes. Under large overlap, the scores are concentrated near the center and the treatment-specific effective sample sizes remain large. Under small overlap, the distribution is wider, more scores approach the boundaries, and the effective sample sizes decrease.
\end{minipage}
\par
\end{figure}

To characterize the treatment-assignment mechanisms, we generated a separate Monte Carlo population of size \(10^6\) under each overlap regime. Figure~\ref{fig:sim_overlap} displays the true propensity-score distributions and the induced treatment allocation. Under large overlap, 45.2\% of the population received treatment and 54.8\% received control; under small overlap, the corresponding proportions were 52.6\% and 47.4\%.

We also calculated the inverse-probability-weighted effective sample size within each treatment group,
\[
\operatorname{ESS}
=
\frac{(\sum_i w_i)^2}{\sum_i w_i^2},
\]
where \(w_i=1/e(W_i)\) among treated individuals and \(w_i=1/\{1-e(W_i)\}\) among controls. This unequal-weight effective-sample-size approximation is used here as an IPW diagnostic \citep{Kish1965SurveySampling,AustinStuart2015IPTW}. Under large overlap, the effective sample sizes were 435,086 for \(A=1\) and 533,240 for \(A=0\). Under small overlap, they decreased to 72,499 and 286,501, respectively. The treatment allocation was therefore approximately balanced in both regimes, although the small-overlap mechanism produced substantial weight concentration.

Continuous potential outcomes were generated as
\[
Y_0
=
\beta^\top(W_1,W_2,W_3,W_7,W_8,W_9)^\top
+
\varepsilon,
\qquad
\varepsilon\sim N(0,1),
\]
with
\[
\beta=(0.8,-0.4,0.6,-0.3,0.5,-0.2)^\top.
\]
Under the homogeneous treatment-effect scenario,
\[
Y_1=Y_0+\delta,
\qquad
\delta=0.5,
\]
whereas under the heterogeneous treatment-effect scenario,
\[
Y_1
=
Y_0+\delta+0.5W_1+0.3W_3+0.2W_1W_3.
\]
The observed continuous outcome was \(Y_{\mathrm{cont}}=AY_1+(1-A)Y_0\).

The ordinal endpoint was obtained by discretizing \(Y_{\mathrm{cont}}\) into four ordered categories using fixed cutpoints. For each scenario, the cutpoints were computed from a Monte Carlo population of size \(10^6\), using the combined distribution of the continuous potential outcomes under treatment and control. The same cutpoints were applied in every replication. The true treatment-specific ordinal distributions and DOOR probability were computed from the same Monte Carlo population.

\subsection{Working-model specification and nuisance learning}
\label{subsec:sim_working_models}

Two covariate representations were evaluated. In the W-oracle analysis, the transformed covariates \(W=\phi(X)\), which determine the treatment and outcome mechanisms, were supplied to the nuisance models. In the X-working analysis, the models received the original covariates rather than the nonlinear transformations. The propensity-score model used \((X_1,\ldots,X_6)\), whereas the outcome hazard models used \((X_1,X_2,X_3,X_7,X_8,X_9)\).

For IPW, we also considered an outcome-oriented propensity-score specification motivated by variable-selection considerations \citep{Brookhart2006}. This specification, denoted OptPS, used \((X_1,X_2,X_3,X_7,X_8,X_9)\) in the X-working analysis, with the analogous \(W\)-covariates used in the W-oracle analysis.

The GLM strategy used main-effects logistic regressions. The Super Learner library included a main-effects GLM, stepwise GLM, a GLM with pairwise interactions, elastic-net regression, generalized additive models, and random forests. Ensemble weights were estimated by 10-fold cross-validation. The library was chosen to include stable parametric learners, regularized regression, smooth nonlinear models, and tree-based interactions \citep{Cannas2019MLPS,Phillips2023SL}.

G-computation used the ordinal hazard model for point estimation; IPW used the propensity-score model. AIPW and TMLE used both nuisance functions. All fitted probabilities were truncated to \([0.001,0.999]\).

\subsection{Treatment-allocation balance}
\label{subsec:sim_imbalance}

Treatment-allocation balance was included as a design factor in the
inferential simulations. The balanced scenarios used the original
propensity-score mechanisms in Section~\ref{subsec:sim_dgm}. To construct
the corresponding imbalanced scenarios while preserving the original
covariate-dependent ordering of treatment probabilities, we shifted the
propensity-score intercept:
\[
e_{\rho}(W)
=
\operatorname{expit}\{\eta(W)+c_{\rho}\}.
\]
For each overlap mechanism, \(c_{\rho}\) was calibrated in a Monte Carlo
population of size \(10^6\) to satisfy
\[
\mathbb{E}\{e_{\rho}(W)\}=\rho,
\qquad
\rho=0.25.
\]
Treatment was then generated as
\[
A\sim\operatorname{Bernoulli}\{e_{\rho}(W)\}.
\]
The imbalanced mechanism therefore produced approximately 25\% treated
and 75\% control observations while retaining the slope coefficients of
the corresponding balanced propensity-score mechanism. Combining the two
treatment-allocation settings with the two overlap regimes and two
treatment-effect structures yielded eight inferential scenarios.

\subsection{Inferential estimator comparison}
\label{subsec:sim_crossfit}

The inferential simulations were restricted to the X-working setting with \(n=2000\). AIPW-SL, TMLE-SL, CVAIPW-SL, and CVTMLE-SL were evaluated across the four balanced scenarios and their four imbalanced-treatment counterparts. The cross-fitted estimators used \(V=10\) folds and otherwise followed the procedures defined in Section~\ref{subsec:crossfit}. This design directly compared each full-sample estimator with its cross-fitted counterpart.

\subsection{Evaluation metrics}
\label{subsec:sim_metrics}

Let \(\hat\psi^{(r)}\) denote the DOOR estimate from replication \(r=1,\ldots,R\), and let \(\psi_0\) denote the Monte Carlo truth. For the point-estimation analyses, we report the Monte Carlo mean estimate, absolute bias, percent bias, root mean squared error, empirical standard error, and the maximum-bias summaries for the cumulative and category probabilities defined below. For the inferential analyses, we additionally report the Monte Carlo mean influence-function-based standard error, coverage of the nominal \(95\%\) transformed Wald interval, and the Monte Carlo average lower and upper confidence limits.

Percent bias, with its sign retained, was calculated as
\[
100
\left\{
\frac{R^{-1}\sum_{r=1}^R\hat\psi^{(r)}-\psi_0}{\psi_0}
\right\}.
\]
To assess recovery of the underlying ordinal outcome distributions, we additionally report
\[
\theta_{\max}
=
\max_{a,k}
\left|
R^{-1}\sum_{r=1}^R\hat\theta_k^{a,(r)}-\theta_k^a
\right|
\]
and
\[
p_{\max}
=
\max_{a,k}
\left|
R^{-1}\sum_{r=1}^R\hat p_k^{a,(r)}-p_k^a
\right|.
\]

\subsection{Simulation results}
\label{subsec:sim_results}

We first evaluated point-estimation performance across all candidate estimators before conducting a focused inferential comparison among the doubly robust estimators. The complete point-estimation results are reported in Supplementary Tables S1--S4. Supplementary Table S1 reports the W-oracle results with \(n=2000\), Supplementary Table S2 reports the X-working results with \(n=2000\), Supplementary Table S3 reports the W-oracle results with \(n=500\), and Supplementary Table S4 reports the X-working results with \(n=500\).

The W-oracle analyses were used as a benchmark for the setting in which the nuisance models were supplied the transformed covariates used in the data-generating mechanism. In the large-sample W-oracle scenarios, most outcome-regression-based and doubly robust estimators were centered close to the true DOOR probability. This pattern was expected, because the GLM working models were already close to the data-generating representation. The finite-sample W-oracle results showed the same qualitative pattern, although limited overlap and smaller risk sets increased variability. Weighting-based estimators were the most sensitive to limited overlap, particularly IPW with Super Learner propensity-score estimation.

\begin{figure}[!tbp]
\centering
\makebox[\linewidth][c]{%
\includegraphics[width=0.98\linewidth]{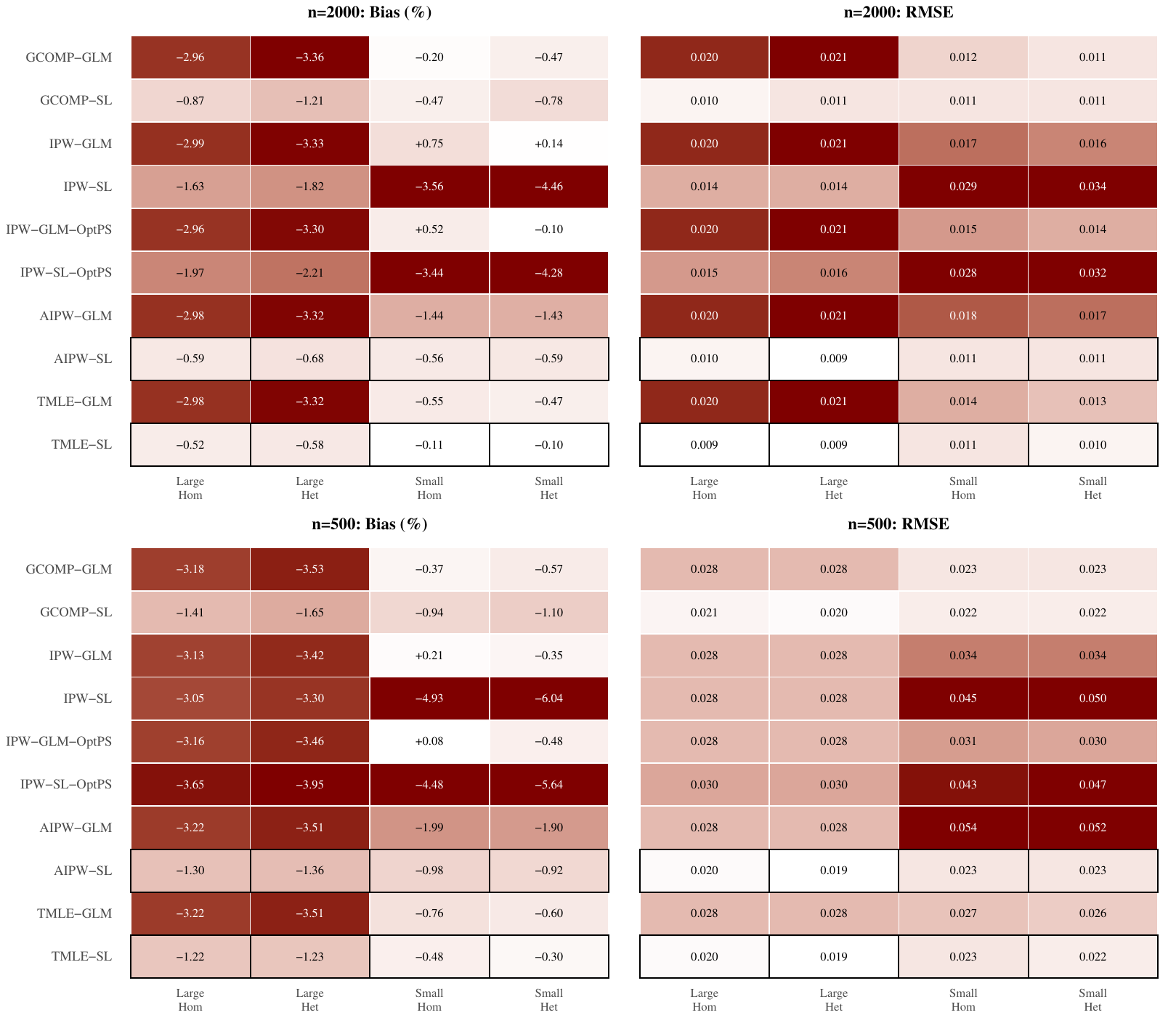}
}
\caption{Point-estimation operating characteristics in the X-working-model simulations. Cells report bias percentage and RMSE for the DOOR probability. Darker shading indicates larger absolute bias in the bias panels and larger RMSE in the RMSE panels. Color intensity is scaled separately within each sample-size panel and metric to improve within-panel contrast; numerical values should be used for cross-panel comparisons. Black outlines highlight the Super-Learner-based doubly robust estimators.}
\label{fig:point_perf_x}
\par\smallskip
\begin{minipage}{0.92\linewidth}
\footnotesize
\textit{Alt text:} Heat maps compare DOOR-probability bias and root mean squared error across estimators, sample sizes, overlap regimes, and homogeneous or heterogeneous treatment effects. TMLE-SL shows the most consistently favorable performance, with AIPW-SL ranking second overall, whereas weighting-based estimators are more sensitive to small overlap.
\end{minipage}
\par
\end{figure}

We then considered the X-working analyses, where the nuisance models received the original baseline covariates rather than the nonlinear transformations used to generate treatment and outcome. This setting is more representative of practical analyses, where the analyst observes baseline variables but not the true nonlinear data-generating representation. Figure~\ref{fig:point_perf_x} summarizes the scalar DOOR point-estimation performance in the X-working simulations by bias percentage and RMSE.

In the X-working analyses, main-effects GLM estimators showed appreciable bias under large overlap, reflecting working-model misspecification. Super Learner reduced bias and RMSE for G-computation, AIPW, and TMLE. Across the \(n=2000\) X-working scenarios, TMLE-SL had the most consistently favorable scalar DOOR performance, with small bias and low RMSE across all overlap and treatment-effect settings. AIPW-SL was close behind. G-computation with Super Learner was also competitive for the scalar DOOR probability in several scenarios, particularly under small overlap, although it did not uniformly match TMLE-SL.

\begin{figure}[!tbp]
\centering
\makebox[\linewidth][c]{%
\includegraphics[width=0.98\linewidth]{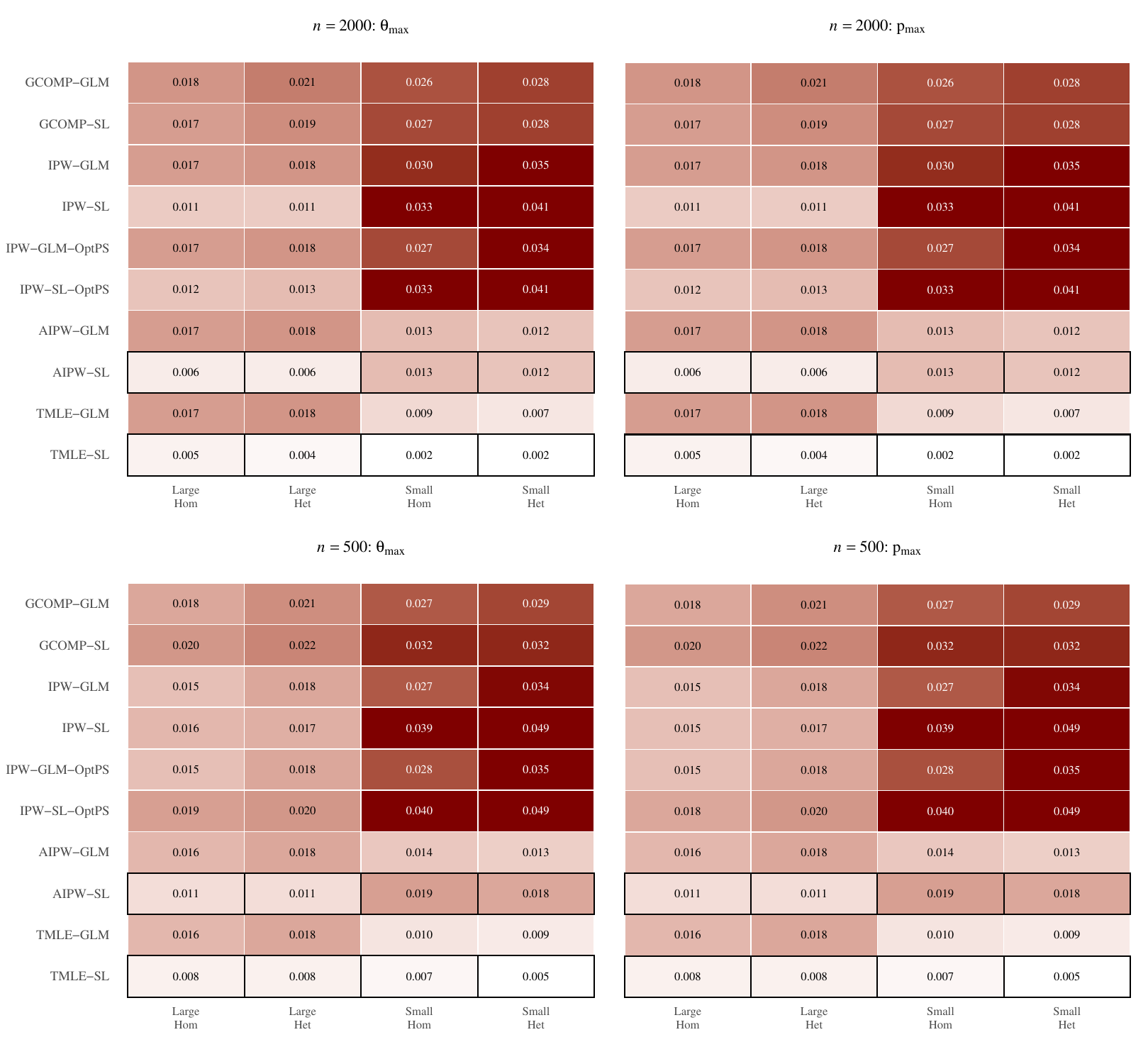}
}
\caption{Maximum biases in the estimated ordinal outcome distributions in the X-working-model simulations. Cells report \(\theta_{\max}\), the largest absolute Monte Carlo mean bias across the cumulative probabilities under the two treatment strategies, and \(p_{\max}\), the corresponding quantity for the category probabilities. Darker shading indicates larger bias. Color intensity is scaled separately within each sample-size panel and metric to improve within-panel contrast; numerical values should be used for cross-panel comparisons. Black outlines highlight the Super-Learner-based doubly robust estimators.}
\label{fig:component_bias_x}
\par\smallskip
\begin{minipage}{0.92\linewidth}
\footnotesize
\textit{Alt text:} Heat maps compare the largest absolute biases in the cumulative and category probabilities under the two treatment strategies. TMLE-SL has the smallest and most consistent errors, AIPW-SL generally ranks second, and GCOMP-SL can show larger distributional errors despite competitive DOOR-scale performance.
\end{minipage}
\par
\end{figure}

To determine whether the scalar DOOR results reflected accurate recovery of the underlying ordinal outcome distributions, we examined the largest absolute Monte Carlo mean biases across the cumulative and category probabilities under the two treatment strategies. Figure~\ref{fig:component_bias_x} reports \(\theta_{\max}\) for the cumulative probabilities \(\theta_k^a=P(Y^a\le k)\) and \(p_{\max}\) for the category probabilities \(p_k^a=P(Y^a=k)\).

These maximum-bias summaries revealed a distinction that was not fully visible from the scalar DOOR probability alone. In the X-working \(n=2000\) small-overlap scenarios, GCOMP-SL had \(\theta_{\max}\) and \(p_{\max}\) values around 0.027--0.028, whereas TMLE-SL had values around 0.002 and AIPW-SL had values around 0.012--0.013. Thus, even when GCOMP-SL had competitive DOOR-scale bias and RMSE, it recovered the underlying marginal ordinal outcome distributions less accurately.

This pattern is compatible with the bilinear structure of the DOOR probability. The DOOR map aggregates the marginal ordinal outcome distributions under the two treatment strategies through \(p^{1\top}Mp^0\). Errors in individual category probabilities can therefore partially offset when mapped to the scalar pairwise probability. A method may consequently have small DOOR-scale bias while recovering portions of the underlying ordinal outcome distributions less accurately. Because the proposed framework estimates both marginal ordinal outcome distributions before applying the DOOR map, their accurate recovery is relevant, particularly when they may also be used for related DOOR-type summaries.

Taken together, the point-estimation results established a clear ordering among the leading non-cross-fitted estimators. TMLE-SL provided the strongest and most consistent performance, with low DOOR-scale bias and RMSE and small maximum biases across the cumulative and category probabilities under the two treatment strategies. AIPW-SL ranked second overall, and these two doubly robust estimators were clearly separated from the remaining methods when the scalar DOOR results and recovery of the underlying ordinal distributions were considered jointly. GCOMP-SL was the only other estimator that approached their scalar DOOR performance in some scenarios, but its larger maximum biases across the cumulative and category probabilities placed it in a lower tier overall. These results motivated the focused inferential analysis of the Super-Learner-based doubly robust estimators.

The inferential analysis was restricted to the X-working \(n=2000\) setting and included three design factors: overlap, treatment-effect heterogeneity, and treatment-allocation balance. We retained the original balanced design and added an imbalanced design to mimic observational settings in which the treated group may be substantially smaller than the control group. This imbalanced design was intended as a practical stress test of unequal treatment prevalence. Cross-fitted versions of AIPW-SL and TMLE-SL were included because cross-fitting separates nuisance-function training from evaluation of the influence-function-based estimating equations. When combined with orthogonal scores, this construction reduces the impact of regularization and overfitting bias and supports asymptotically valid inference with flexible nuisance learning \citep{Chernozhukov2018DML}. Prior causal inference work has also shown that machine-learning-based singly robust estimators may perform poorly, whereas doubly robust estimators combined with sample splitting can improve bias and coverage \citep{Naimi2023MLCausal}. Related CVTMLE work has shown that cross-fitting can improve coverage without materially increasing bias, particularly under small samples, near-positivity violations, complex machine-learning nuisance estimators, or unbalanced exposure prevalence \citep{Smith2025CVTMLE}.

\begin{table}[h]
\centering
\caption{Inferential operating characteristics for the Super-Learner-based doubly robust estimators in the X-working, large-sample setting (\(n=2000\), 1000 replications). Results are shown for TMLE-SL, CVTMLE-SL, AIPW-SL, and CVAIPW-SL across treatment-allocation, overlap, and treatment-effect heterogeneity scenarios. Bias, RMSE, empirical standard error (Emp. SE), and influence-function-based standard error (IF SE) are reported on the DOOR probability scale. Coverage is based on transformed Wald confidence intervals; Lower and Upper are the Monte Carlo average interval limits after transformation back to the DOOR probability scale. The quantities \(\theta_{\max}\) and \(p_{\max}\) are the largest absolute Monte Carlo mean biases across the cumulative and category probabilities, respectively, under the two treatment strategies.}
\label{tab:cv_formal_1000rep}
\resizebox{\textwidth}{!}{%
\begin{tabular}{llrrrrrrrrrr}
\toprule
Scenario & Estimator & Abs. bias & Bias (\%) & RMSE & Emp. SE & IF SE & Coverage (\%) & Lower & Upper & \(\theta_{\max}\) & \(p_{\max}\) \\
\midrule
\multirow{4}{*}{\parbox{4.2cm}{\centering \textbf{Balanced design,\\ Large overlap,\\ Homogeneous effect}\\ \(\boldsymbol{\psi}_0=0.584\)}}& TMLE--SL & 0.003 & -0.53 & 0.009 & 0.009 & 0.009 & 92.00 & 0.564 & 0.598 & 0.005 & 0.005 \\
& CVTMLE--SL & 0.003 & -0.47 & 0.009 & 0.009 & 0.009 & 93.90 & 0.563 & 0.599 & 0.004 & 0.004 \\
& AIPW--SL & 0.003 & -0.59 & 0.010 & 0.009 & 0.009 & 91.90 & 0.564 & 0.597 & 0.006 & 0.006 \\
& CVAIPW--SL & 0.003 & -0.48 & 0.009 & 0.009 & 0.009 & 94.00 & 0.563 & 0.599 & 0.004 & 0.004 \\
\midrule

\multirow{4}{*}{\parbox{4.2cm}{\centering \textbf{Balanced design,\\ Large overlap,\\ Heterogeneous effect}\\ \(\boldsymbol{\psi}_0=0.560\)}}& TMLE--SL & 0.003 & -0.58 & 0.009 & 0.008 & 0.008 & 92.20 & 0.541 & 0.573 & 0.004 & 0.004 \\
& CVTMLE--SL & 0.003 & -0.51 & 0.009 & 0.008 & 0.009 & 94.30 & 0.540 & 0.575 & 0.004 & 0.004 \\
& AIPW--SL & 0.004 & -0.68 & 0.009 & 0.008 & 0.008 & 90.90 & 0.540 & 0.573 & 0.006 & 0.006 \\
& CVAIPW--SL & 0.003 & -0.52 & 0.009 & 0.008 & 0.009 & 94.30 & 0.540 & 0.575 & 0.004 & 0.004 \\
\midrule

\multirow{4}{*}{\parbox{4.2cm}{\centering \textbf{Balanced design,\\ Small overlap,\\ Homogeneous effect}\\ \(\boldsymbol{\psi}_0=0.584\)}}& TMLE--SL & 0.001 & -0.10 & 0.011 & 0.011 & 0.010 & 92.20 & 0.565 & 0.602 & 0.002 & 0.002 \\
& CVTMLE--SL & 0.001 & -0.12 & 0.011 & 0.011 & 0.011 & 95.40 & 0.562 & 0.605 & 0.003 & 0.003 \\
& AIPW--SL & 0.003 & -0.57 & 0.011 & 0.011 & 0.010 & 92.00 & 0.561 & 0.600 & 0.013 & 0.013 \\
& CVAIPW--SL & 0.003 & -0.53 & 0.012 & 0.011 & 0.011 & 95.60 & 0.558 & 0.603 & 0.009 & 0.009 \\
\midrule

\multirow{4}{*}{\parbox{4.2cm}{\centering \textbf{Balanced design,\\ Small overlap,\\ Heterogeneous effect}\\ \(\boldsymbol{\psi}_0=0.560\)}}& TMLE--SL & 0.001 & -0.10 & 0.010 & 0.010 & 0.009 & 92.40 & 0.542 & 0.578 & 0.002 & 0.002 \\
& CVTMLE--SL & 0.001 & -0.10 & 0.010 & 0.010 & 0.011 & 95.80 & 0.539 & 0.581 & 0.003 & 0.003 \\
& AIPW--SL & 0.003 & -0.58 & 0.011 & 0.010 & 0.009 & 91.50 & 0.539 & 0.575 & 0.012 & 0.012 \\
& CVAIPW--SL & 0.003 & -0.46 & 0.011 & 0.010 & 0.011 & 95.50 & 0.536 & 0.579 & 0.007 & 0.007 \\
\midrule

\multirow{4}{*}{\parbox{4.2cm}{\centering \textbf{Imbalanced design,\\ Large overlap,\\ Homogeneous effect}\\ \(\boldsymbol{\psi}_0=0.584\)}}& TMLE--SL & 0.003 & -0.55 & 0.011 & 0.010 & 0.009 & 90.70 & 0.562 & 0.599 & 0.006 & 0.006 \\
& CVTMLE--SL & 0.003 & -0.48 & 0.011 & 0.010 & 0.011 & 94.20 & 0.560 & 0.602 & 0.005 & 0.005 \\
& AIPW--SL & 0.004 & -0.76 & 0.011 & 0.010 & 0.010 & 89.00 & 0.561 & 0.598 & 0.009 & 0.009 \\
& CVAIPW--SL & 0.003 & -0.48 & 0.011 & 0.010 & 0.011 & 94.10 & 0.560 & 0.602 & 0.005 & 0.005 \\
\midrule

\multirow{4}{*}{\parbox{4.2cm}{\centering \textbf{Imbalanced design,\\ Large overlap,\\ Heterogeneous effect}\\ \(\boldsymbol{\psi}_0=0.560\)}}& TMLE--SL & 0.004 & -0.65 & 0.011 & 0.010 & 0.009 & 89.50 & 0.539 & 0.574 & 0.005 & 0.005 \\
& CVTMLE--SL & 0.003 & -0.58 & 0.010 & 0.010 & 0.010 & 94.30 & 0.537 & 0.577 & 0.005 & 0.005 \\
& AIPW--SL & 0.005 & -0.92 & 0.011 & 0.010 & 0.009 & 87.50 & 0.537 & 0.573 & 0.009 & 0.009 \\
& CVAIPW--SL & 0.003 & -0.58 & 0.011 & 0.010 & 0.010 & 94.40 & 0.537 & 0.577 & 0.005 & 0.005 \\
\midrule

\multirow{4}{*}{\parbox{4.2cm}{\centering \textbf{Imbalanced design,\\ Small overlap,\\ Homogeneous effect}\\ \(\boldsymbol{\psi}_0=0.584\)}}& TMLE--SL & 0.001 & -0.24 & 0.016 & 0.016 & 0.010 & 79.30 & 0.562 & 0.603 & 0.004 & 0.004 \\
& CVTMLE--SL & 0.003 & -0.55 & 0.016 & 0.015 & 0.013 & 89.70 & 0.555 & 0.607 & 0.007 & 0.007 \\
& AIPW--SL & 0.009 & -1.58 & 0.017 & 0.014 & 0.011 & 76.10 & 0.554 & 0.596 & 0.023 & 0.023 \\
& CVAIPW--SL & 0.006 & -1.01 & 0.016 & 0.015 & 0.014 & 91.10 & 0.551 & 0.605 & 0.015 & 0.015 \\
\midrule

\multirow{4}{*}{\parbox{4.2cm}{\centering \textbf{Imbalanced design,\\ Small overlap,\\ Heterogeneous effect}\\ \(\boldsymbol{\psi}_0=0.560\)}}& TMLE--SL & 0.000 & -0.06 & 0.014 & 0.014 & 0.010 & 82.70 & 0.540 & 0.580 & 0.003 & 0.005 \\
& CVTMLE--SL & 0.003 & -0.47 & 0.014 & 0.014 & 0.013 & 92.80 & 0.533 & 0.582 & 0.005 & 0.005 \\
& AIPW--SL & 0.009 & -1.68 & 0.016 & 0.013 & 0.010 & 76.40 & 0.531 & 0.571 & 0.022 & 0.022 \\
& CVAIPW--SL & 0.006 & -0.99 & 0.015 & 0.013 & 0.013 & 93.10 & 0.529 & 0.580 & 0.013 & 0.013 \\
\bottomrule
\end{tabular}%
}
\end{table}

Table~\ref{tab:cv_formal_1000rep} reports the inferential operating characteristics for TMLE-SL, CVTMLE-SL, AIPW-SL, and CVAIPW-SL. In the balanced scenarios, cross-fitting generally preserved the small bias and RMSE of the non-cross-fitted estimators while moving coverage closer to the nominal level. This was most apparent in the small-overlap scenarios. For example, in the balanced small-overlap homogeneous scenario, TMLE-SL had 92.20\% coverage, whereas CVTMLE-SL had 95.40\% coverage; AIPW-SL had 92.00\% coverage, whereas CVAIPW-SL had 95.60\% coverage. Similar improvements were observed in the balanced small-overlap heterogeneous scenario.

The imbalanced scenarios were more difficult, particularly when treatment allocation imbalance was combined with small overlap. In these two scenarios, the non-cross-fitted estimators showed clear undercoverage. TMLE-SL had coverage of 79.30\% and 82.70\%, and AIPW-SL had coverage of 76.10\% and 76.40\%, in the homogeneous and heterogeneous settings, respectively. Cross-fitting substantially improved coverage. CVTMLE-SL increased coverage to 89.70\% and 92.80\%, and CVAIPW-SL increased coverage to 91.10\% and 93.10\%.

The coverage gains after cross-fitting were accompanied by closer agreement between the influence-function-based and empirical standard errors. This pattern was most apparent in the imbalanced small-overlap scenarios. In those scenarios, the influence-function-based standard error was clearly smaller than the empirical standard error for both TMLE-SL and AIPW-SL. After cross-fitting, the two standard errors were more closely aligned, together with improved transformed-Wald coverage.

Between the two cross-fitted estimators, CVAIPW-SL attained slightly higher coverage in some of the most difficult scenarios, particularly when treatment-allocation imbalance was combined with small overlap. However, CVTMLE-SL generally had equal or lower RMSE, smaller absolute DOOR-scale bias, and smaller maximum biases across the cumulative and category probabilities under the two treatment strategies. Across the inferential scenarios, CVTMLE-SL therefore showed the strongest and most consistent performance among the cross-fitted estimators evaluated.

\section{Real Data Application}
\label{sec:appl}

\subsection{Study population, treatment strategies, and DOOR endpoint}
\label{subsec:rwd_data}

We analyzed an updated dataset from the Multi-Drug Resistant Organism (MDRO) network, a collection of observational studies conducted within the Antibacterial Resistance Leadership Group (ARLG). The network included patients with carbapenem-resistant \textit{Acinetobacter baumannii} from SNAP \citep{c29}, carbapenem-resistant \textit{Pseudomonas aeruginosa} from POP \citep{c30}, and carbapenem-resistant \textit{Enterobacterales} from CRACKLE \citep{c40}. Data from the three studies were combined to compare the benefit-risk profiles of initial monotherapy and combination therapy recorded on the day of culture collection and the following day. Concurrent receipt of two or more eligible antibacterial agents during this period was classified as combination therapy; otherwise, treatment was classified as monotherapy. A related analysis of the MDRO network using parametric nuisance models was reported by \citet{Shu2026}.

The analytic dataset used for the estimates reported below included 1535 patients, of whom 1145 received monotherapy and 390 received combination therapy. The combination-to-monotherapy ratio was therefore approximately \(1:2.94\), close to the \(1:3\) allocation examined in the imbalanced inferential simulations. The four-category DOOR endpoint was ordered from most to least desirable as follows: alive without a qualifying event; alive with one qualifying event; alive with two or three qualifying events; and death. Qualifying events included absence of clinical response, infectious complications, and nonfatal serious adverse events. For estimation, the categories were coded so that larger values represented more desirable outcomes, consistent with the notation used throughout the article.

Covariates used for adjustment included geographic location, categorized as the United States, South America, or other; study; Pitt bacteremia score, entered continuously; age-adjusted Charlson Comorbidity Index; and infection site. Table~\ref{tab:rwd_descriptive} summarizes these variables together with age and intensive care unit status as additional clinical descriptors.

\begin{table}[!tbp]
\centering
\caption{Patient characteristics and observed DOOR outcomes by treatment strategy in the MDRO application.}
\label{tab:rwd_descriptive}
\small
\begin{tabular}{p{0.42\linewidth}ccc}
\toprule
Characteristic
& \makecell{Monotherapy\\(\(n=1145\))}
& \makecell{Combination therapy\\(\(n=390\))}
& \makecell{\(P\)\\value} \\
\midrule
\textit{Observed DOOR category, \(n\) (\%)} & & & \(<0.001\) \\
\quad Alive without a qualifying event
& 451 (39) & 106 (27) & \\
\quad Alive with one qualifying event
& 242 (21) & 84 (22) & \\
\quad Alive with two or three qualifying events
& 205 (18) & 81 (21) & \\
\quad Death
& 247 (22) & 119 (31) & \\[2pt]

\textit{Geographic location, \(n\) (\%)} & & & 0.19 \\
\quad United States
& 873 (76) & 287 (74) & \\
\quad South America
& 188 (16) & 79 (20) & \\
\quad Other
& 84 (7.3) & 24 (6.2) & \\[2pt]

\textit{Study, \(n\) (\%)} & & & \(<0.001\) \\
\quad CRACKLE (CRE)
& 642 (56) & 224 (57) & \\
\quad POP (CRPA)
& 359 (31) & 92 (24) & \\
\quad SNAP (CRAB)
& 144 (13) & 74 (19) & \\[2pt]

Pitt bacteremia score, median (IQR)
& 3.00 (2.00--6.00) & 4.00 (2.00--6.00) & \(<0.001\) \\
Age-adjusted Charlson Comorbidity Index, median (IQR)
& 2.00 (1.00--4.00) & 2.00 (1.00--4.00) & 0.48 \\
Age at culture collection, median (IQR)
& 62 (49--72) & 61 (42--69) & 0.014 \\
ICU, \(n\) (\%)
& 676 (59) & 274 (70) & \(<0.001\) \\[2pt]

\textit{Infection site, \(n\) (\%)} & & & \(<0.001\) \\
\quad Blood
& 346 (30) & 192 (49) & \\
\quad Respiratory
& 371 (32) & 116 (30) & \\
\quad Urine
& 164 (14) & 34 (8.7) & \\
\quad Other
& 264 (23) & 48 (12) & \\
\bottomrule
\end{tabular}
\par\smallskip
\begin{minipage}{0.96\textwidth}
\footnotesize
\textit{Note:} \(P\) values compare the observed treatment groups and were calculated using Pearson's chi-squared test for categorical variables and the Kruskal--Wallis rank-sum test for continuous variables; smaller values indicate stronger evidence of an unadjusted group difference. CRE, carbapenem-resistant \textit{Enterobacterales}; CRPA, carbapenem-resistant \textit{Pseudomonas aeruginosa}; CRAB, carbapenem-resistant \textit{Acinetobacter baumannii}; ICU, intensive care unit; IQR, interquartile range.
\end{minipage}
\par
\end{table}

Patients receiving combination therapy had a less favorable observed DOOR distribution, including a higher proportion of death. They also had higher Pitt bacteremia scores and greater proportions of intensive care unit admission and bloodstream infection. These unadjusted differences provide clinical context for the covariate-adjusted analysis.

\subsection{Estimation and results}
\label{subsec:rwd_results}

Combination therapy was coded as \(A=1\) and monotherapy as \(A=0\). A DOOR probability below 0.5 favors monotherapy, a value above 0.5 favors combination therapy, and 0.5 represents no overall advantage. We carried forward the four Super-Learner-based doubly robust estimators evaluated in the inferential simulations: AIPW-SL, TMLE-SL, CVAIPW-SL, and CVTMLE-SL. AIPW-SL and TMLE-SL used full-sample nuisance estimates. For CVAIPW-SL and CVTMLE-SL, the propensity score and ordinal hazards were estimated using 10-fold cross-fitting, so each individual received nuisance predictions from models fitted without that individual's fold. For CVTMLE-SL, the out-of-fold predictions were stacked across folds, after which a single hazard-scale targeting update was performed.

All four adjusted estimators used the same six-candidate Super Learner library as in the simulations: main-effects GLM, stepwise GLM, GLM with pairwise interactions, elastic net, generalized additive models, and random forests. Fitted probabilities were bounded to \([0.001,0.999]\). Inference followed the common efficient-influence-function-based framework developed in Section~\ref{subsec:inference}. Their standard errors were calculated from the estimated efficient influence functions, and their \(95\%\) confidence intervals were constructed on the inverse-hyperbolic-tangent scale and transformed back to the probability scale. For the crude analysis, the standard error and \(95\%\) confidence interval were calculated using the Wald-type procedure described in the Supplementary Material of \citet{Hamasaki2025}.

\begin{table}[h]
\centering
\caption{Crude and covariate-adjusted DOOR probability estimates in the MDRO application.}
\label{tab:rwd_results}
\small
\begin{tabular}{lccc}
\toprule
Analysis & Estimate & Standard error & 95\% confidence interval \\
\midrule
Crude (unadjusted) & 0.42236 & 0.01600 & (0.39135, 0.45400) \\
AIPW--SL           & 0.47074 & 0.01467 & (0.44213, 0.49956) \\
TMLE--SL           & 0.47100 & 0.01465 & (0.44241, 0.49978) \\
\addlinespace[2pt]
CVAIPW--SL         & 0.46945 & 0.01577 & (0.43870, 0.50044) \\
CVTMLE--SL         & 0.46901 & 0.01576 & (0.43828, 0.49997) \\
\bottomrule
\end{tabular}
\par\smallskip
\begin{minipage}{0.92\linewidth}
\footnotesize
\textit{Note:} Estimated values below 0.5 favor monotherapy. AIPW, augmented inverse probability weighting; CVAIPW, cross-fitted augmented inverse probability weighting; TMLE, targeted maximum likelihood estimation; CVTMLE, cross-fitted targeted maximum likelihood estimation; SL, Super Learner.
\end{minipage}
\par
\end{table}

\begin{figure}[!tbp]
\centering
\includegraphics[width=0.94\linewidth]{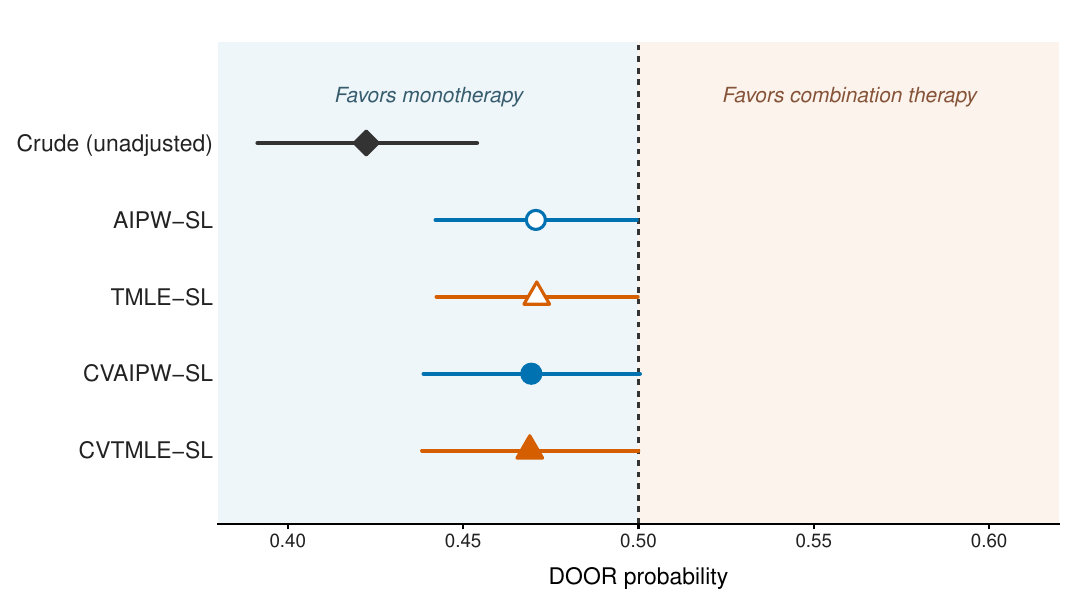}
\caption{Crude and covariate-adjusted DOOR probability estimates in the MDRO application. The full-sample estimators are shown before their 10-fold cross-fitted counterparts. Points denote estimates and horizontal bars denote \(95\%\) confidence intervals. The dashed line marks 0.5; the shaded regions indicate values favoring monotherapy or combination therapy.}
\label{fig:rwd_forest}
\par\smallskip
\begin{minipage}{0.92\linewidth}
\footnotesize
\textit{Alt text:} Forest plot of the crude and four covariate-adjusted DOOR probability estimates. The crude estimate favors monotherapy. All four adjusted estimates are closer to 0.5, and the cross-fitted estimates have slightly wider confidence intervals. Values below 0.5 favor monotherapy, whereas values above 0.5 favor combination therapy.
\end{minipage}
\par
\end{figure}

Table~\ref{tab:rwd_results} reports the numerical results, and Figure~\ref{fig:rwd_forest} displays the same estimates graphically. The crude DOOR probability was 0.422, indicating a more favorable observed outcome distribution under monotherapy before adjustment. The four adjusted estimates ranged from 0.469 to 0.471, representing a shift of approximately 0.047--0.049 toward the null value of 0.5. Thus, the apparent monotherapy advantage was substantially attenuated after adjustment for measured baseline covariates.

The non-cross-fitted and cross-fitted implementations produced similar point estimates. The influence-function-based standard errors were slightly larger for CVAIPW--SL and CVTMLE--SL than for their non-cross-fitted counterparts, consistent with the pattern observed in the inferential simulations. In particular, CVTMLE--SL yielded an estimated DOOR probability of 0.469 with a \(95\%\) confidence interval of \(0.438\) to \(0.500\). The adjusted confidence intervals reached or closely approached 0.5, providing substantially weaker evidence of an overall monotherapy advantage than the crude comparison.

Covariate adjustment does not change the construction or clinical interpretation of the DOOR endpoint. Instead, it estimates the same pairwise benefit-risk probability after standardizing the treatment strategies to a common distribution of measured baseline covariates. Because the MDRO data are observational, the causal interpretation relies on consistency, no interference, conditional exchangeability, and positivity.

\section{Discussion}
\label{sec:disc}

We developed a unified causal inference framework for estimation and inference on the population-level DOOR probability. The framework represents the DOOR probability as a bilinear functional of the marginal ordinal outcome distributions under the two treatment strategies, estimates the conditional ordinal distributions through sequential risk-set hazards, and propagates uncertainty through the efficient influence function and functional delta method. This organization is modular: nuisance learners may be replaced without changing the estimand, and additional doubly robust estimators or cross-fitting schemes can be incorporated once their estimating steps and regularity conditions are established. The present study used a six-candidate Super Learner library and a pooled-targeting implementation of CVTMLE. In the point-estimation simulations, TMLE-SL provided the strongest and most consistent performance, with AIPW-SL ranking second overall. The results also showed that a small scalar DOOR error did not always imply equally accurate recovery of the underlying ordinal outcome distributions. In the inferential simulations, cross-fitting generally improved agreement between influence-function-based and empirical standard errors and moved confidence-interval coverage toward the nominal level. CVTMLE-SL provided the strongest overall performance among the methods evaluated, although modest undercoverage remained in the most difficult settings.

The remaining undercoverage is consistent with recognized finite-sample difficulties in machine-learning-based inference. Related CVTMLE research has identified data sparsity or small samples, near-positivity violations associated with unbalanced exposure prevalence or extrapolation across poorly supported treatment--covariate strata, and highly data-adaptive nuisance learners as settings in which standard TMLE may under-cover \citep{Smith2025CVTMLE}. Several of these features occurred together in our most difficult scenarios. Unequal treatment allocation and limited overlap reduced the information available in one treatment group; the sequential ordinal representation further divided the observations by treatment strategy and successive risk sets; and the Super Learner library included a random-forest candidate, although we did not isolate the contribution of individual learners. In the heterogeneous small-overlap scenarios, \(W_1\) and \(W_3\) affected both treatment assignment and treatment-effect heterogeneity, so some treatment-specific outcome predictions may also have relied more heavily on weakly supported covariate regions. These mechanisms provide plausible explanations for the residual undercoverage, but their individual contributions cannot be separated because they were not independently varied or diagnosed. Cross-fitting reduces the impact of overfitting on the estimating equation by separating nuisance-function training from influence-function evaluation, but it cannot add information to treatment-specific covariate regions or ordinal risk sets with few observations.

The MDRO application illustrated the framework in an observational benefit-risk analysis with 390 patients receiving combination therapy and 1145 receiving monotherapy, an allocation ratio of approximately \(1:2.9\). Covariate adjustment moved the estimated DOOR probability from 0.422 to approximately 0.469--0.471, substantially weakening the crude evidence favoring monotherapy. The four adjusted estimators produced similar point estimates, while the cross-fitted estimators had slightly larger influence-function-based standard errors than their non-cross-fitted counterparts. This direction was consistent with the simulation pattern in which cross-fitting reduced underestimation of sampling variability. A single application, however, does not provide the Monte Carlo empirical standard error needed to determine whether the larger standard errors were more accurate. Adjustment substantially weakened the evidence favoring monotherapy, but the results did not establish equivalence between the two treatment strategies.

Several limitations remain. Causal interpretation requires consistency, no interference, conditional exchangeability, and positivity; the observational MDRO analysis may therefore remain subject to residual confounding and practical positivity limitations. Formal propensity-score overlap, extreme-weight, effective-sample-size, and model-sensitivity diagnostics were not available for that application. The sequential risk-set representation may also become unstable when later risk sets contain few observations or events, particularly with rare outcome categories, unequal treatment allocation, or limited overlap. Performance may further depend on the learner library, tuning choices, probability bounds, and computational resources. The simulations considered four ordinal categories and a prespecified range of sample sizes, overlap regimes, treatment-effect structures, and treatment allocations. The observed ordering of the estimators should therefore not be assumed to hold in every DOOR application.

Future work should compare alternative learner libraries, fold numbers, repeated cross-fitting, and fold-specific versus pooled targeting, as well as methods for stabilizing estimation when later ordinal risk sets contain few observations or events. Neural networks and other highly adaptive learners could be evaluated when sufficient information is available. Data-adaptive variable selection also warrants further study, although predictive algorithms alone cannot determine which measured variables are confounders, and adjustment decisions should remain guided by clinical and causal knowledge. Additional simulations should independently vary extrapolation, rare outcome categories, stronger positivity violations, and smaller treatment-specific risk sets. Finally, the DOOR probability is a population-level pairwise estimand and does not identify whether a particular individual would benefit from treatment. CVTMLE-SL should therefore be viewed as the preferred implementation among the methods and settings evaluated here, rather than as a universally optimal estimator, with CVAIPW-SL providing a viable doubly robust alternative.

\section*{Acknowledgments.}
During the preparation of this manuscript, OpenAI ChatGPT and Google Gemini were used for critical review, generation of counterarguments, and language and grammar review, while OpenAI Codex was used as an interactive computational interface to orchestrate the parallel execution of local simulation code and reduce computational turnaround time. The authors retained full intellectual and scientific control over source selection, code development and validation, verification of outputs, scientific interpretation, and final wording, and take full responsibility for the manuscript; no individual-level or restricted real-world data were uploaded to these tools.

\section*{Supplementary Material}

The complete point-estimation simulation results are provided in Supplementary Tables~S1--S4 in the separate Supplementary Materials file accompanying this manuscript.

\section*{Funding}
This work was supported by the National Institute of Allergy and Infectious Diseases of the National Institutes of Health [UM1AI104681]. The content is solely the responsibility of the authors and does not necessarily represent the official views of the National Institutes of Health.

\section*{Conflict of interest.}
None declared.

\section*{Data Availability Statement}

Individual deidentified participant data (and supporting documentation, data dictionaries, and protocol) that underlie the results in this article can be made available to investigators following submission of a plan for data use, approval by the ARLG or designated entity, and execution of required institutional agreements. Provision might be contingent upon the availability of funding for data preparation and deidentification. More information can be found at: \url{https://arlg.org/request-data/}.

\bibliographystyle{biorefs}
\bibliography{bibliography}
\clearpage
\markboth{}{} 

\setcounter{page}{1} 

\section*{Supplementary Materials}
\setcounter{table}{0}
\renewcommand{\thetable}{S\arabic{table}}

Supplementary Tables S1--S4 provide the complete point-estimation simulation results. They are ordered by sample size and working-model input: W-oracle with \(n=2000\), X-working with \(n=2000\), W-oracle with \(n=500\), and X-working with \(n=500\). Each table reports the Monte Carlo mean estimate on the DOOR probability scale, absolute bias, signed percent bias, root mean squared error (RMSE), and empirical standard error. The quantities \(\theta_{\max}\) and \(p_{\max}\) denote the largest absolute Monte Carlo mean biases in the cumulative and category probabilities, respectively, across the two treatment strategies. Inferential operating characteristics are reported in the main manuscript.

\newcommand{\biaszero}{\(<0.001\)}

\begin{table}[H]
\centering
\caption{Point-estimation results for the W-oracle, large-sample setting (\(n=2000\)). The transformed covariates \(W=\phi(X)\), which drive the true treatment and outcome mechanisms, were supplied to the nuisance models.}
\label{tab:supp_w_n2000}
\resizebox{\textwidth}{!}{%
\begin{tabular}{llrrrrrrr}
\toprule
Scenario & Estimator & Mean est. & Abs. bias & Bias (\%) & RMSE & Emp. SE & \(\theta_{\max}\) & \(p_{\max}\) \\
\midrule

\multirow{10}{*}{\parbox{4.2cm}{\centering \textbf{Large overlap,\\ Homogeneous effect}\\ \(\boldsymbol{\psi}_0=0.584\)}}
& GCOMP-GLM      & 0.584 & \biaszero & -0.02 & 0.009 & 0.009 & 0.002 & 0.003 \\
& GCOMP-SL       & 0.583 & \biaszero & -0.14 & 0.009 & 0.009 & 0.004 & 0.004 \\
& IPW-GLM        & 0.584 & \biaszero & -0.03 & 0.010 & 0.010 & 0.001 & 0.002 \\
& IPW-SL         & 0.581 & 0.003 & -0.52 & 0.011 & 0.011 & 0.004 & 0.004 \\
& IPW-GLM-OptPS  & 0.584 & \biaszero & 0.00 & 0.009 & 0.009 & 0.001 & 0.002 \\
& IPW-SL-OptPS   & 0.580 & 0.004 & -0.73 & 0.011 & 0.010 & 0.005 & 0.005 \\
& AIPW-GLM       & 0.584 & \biaszero & 0.01 & 0.009 & 0.009 & 0.001 & 0.002 \\
& AIPW-SL        & 0.584 & \biaszero & -0.07 & 0.009 & 0.009 & 0.002 & 0.002 \\
& TMLE-GLM       & 0.584 & \biaszero & 0.01 & 0.009 & 0.009 & 0.001 & 0.002 \\
& TMLE-SL        & 0.584 & \biaszero & -0.06 & 0.009 & 0.009 & 0.002 & 0.002 \\
\midrule

\multirow{10}{*}{\parbox{4.2cm}{\centering \textbf{Large overlap,\\ Heterogeneous effect}\\ \(\boldsymbol{\psi}_0=0.560\)}}
& GCOMP-GLM      & 0.560 & \biaszero & -0.05 & 0.008 & 0.008 & 0.002 & 0.002 \\
& GCOMP-SL       & 0.559 & 0.001 & -0.20 & 0.008 & 0.008 & 0.004 & 0.004 \\
& IPW-GLM        & 0.560 & \biaszero & -0.05 & 0.009 & 0.009 & 0.001 & 0.001 \\
& IPW-SL         & 0.557 & 0.003 & -0.58 & 0.011 & 0.010 & 0.004 & 0.004 \\
& IPW-GLM-OptPS  & 0.560 & \biaszero & -0.02 & 0.009 & 0.009 & 0.001 & 0.001 \\
& IPW-SL-OptPS   & 0.556 & 0.005 & -0.86 & 0.011 & 0.010 & 0.006 & 0.006 \\
& AIPW-GLM       & 0.560 & \biaszero & -0.01 & 0.008 & 0.008 & 0.001 & 0.001 \\
& AIPW-SL        & 0.560 & \biaszero & -0.10 & 0.008 & 0.008 & 0.002 & 0.002 \\
& TMLE-GLM       & 0.560 & \biaszero & -0.01 & 0.008 & 0.008 & 0.001 & 0.001 \\
& TMLE-SL        & 0.560 & \biaszero & -0.08 & 0.008 & 0.008 & 0.002 & 0.002 \\
\midrule

\multirow{10}{*}{\parbox{4.2cm}{\centering \textbf{Small overlap,\\ Homogeneous effect}\\ \(\boldsymbol{\psi}_0=0.584\)}}
& GCOMP-GLM      & 0.583 & \biaszero & -0.13 & 0.010 & 0.009 & 0.005 & 0.005 \\
& GCOMP-SL       & 0.581 & 0.003 & -0.50 & 0.010 & 0.010 & 0.009 & 0.009 \\
& IPW-GLM        & 0.583 & 0.001 & -0.24 & 0.019 & 0.019 & 0.004 & 0.004 \\
& IPW-SL         & 0.569 & 0.015 & -2.52 & 0.027 & 0.023 & 0.019 & 0.019 \\
& IPW-GLM-OptPS  & 0.582 & 0.003 & -0.44 & 0.016 & 0.016 & 0.007 & 0.007 \\
& IPW-SL-OptPS   & 0.570 & 0.015 & -2.49 & 0.025 & 0.020 & 0.020 & 0.020 \\
& AIPW-GLM       & 0.584 & \biaszero & 0.02 & 0.011 & 0.011 & 0.001 & 0.001 \\
& AIPW-SL        & 0.583 & 0.001 & -0.25 & 0.011 & 0.011 & 0.004 & 0.004 \\
& TMLE-GLM       & 0.582 & 0.002 & -0.33 & 0.017 & 0.017 & 0.004 & 0.004 \\
& TMLE-SL        & 0.582 & 0.002 & -0.40 & 0.012 & 0.012 & 0.005 & 0.005 \\
\midrule

\multirow{10}{*}{\parbox{4.2cm}{\centering \textbf{Small overlap,\\ Heterogeneous effect}\\ \(\boldsymbol{\psi}_0=0.560\)}}
& GCOMP-GLM      & 0.561 & \biaszero & 0.03 & 0.009 & 0.009 & 0.002 & 0.002 \\
& GCOMP-SL       & 0.558 & 0.002 & -0.42 & 0.009 & 0.009 & 0.007 & 0.007 \\
& IPW-GLM        & 0.558 & 0.002 & -0.33 & 0.020 & 0.020 & 0.005 & 0.005 \\
& IPW-SL         & 0.543 & 0.017 & -3.12 & 0.031 & 0.025 & 0.023 & 0.023 \\
& IPW-GLM-OptPS  & 0.557 & 0.003 & -0.57 & 0.017 & 0.016 & 0.007 & 0.007 \\
& IPW-SL-OptPS   & 0.543 & 0.017 & -3.07 & 0.028 & 0.022 & 0.024 & 0.024 \\
& AIPW-GLM       & 0.560 & \biaszero & -0.02 & 0.010 & 0.010 & 0.001 & 0.001 \\
& AIPW-SL        & 0.559 & 0.001 & -0.25 & 0.010 & 0.010 & 0.003 & 0.003 \\
& TMLE-GLM       & 0.559 & 0.002 & -0.32 & 0.021 & 0.021 & 0.003 & 0.003 \\
& TMLE-SL        & 0.558 & 0.002 & -0.43 & 0.011 & 0.011 & 0.005 & 0.005 \\
\bottomrule
\end{tabular}%
}
\end{table}

\begin{table}[H]
\centering
\caption{Point-estimation results for the X-working, large-sample setting (\(n=2000\)). The original baseline covariates \(X\), rather than the nonlinear transformed covariates \(W=\phi(X)\), were supplied to the nuisance models.}
\label{tab:supp_x_n2000}
\resizebox{\textwidth}{!}{%
\begin{tabular}{llrrrrrrr}
\toprule
Scenario & Estimator & Mean est. & Abs. bias & Bias (\%) & RMSE & Emp. SE & \(\theta_{\max}\) & \(p_{\max}\) \\
\midrule

\multirow{10}{*}{\parbox{4.2cm}{\centering \textbf{Large overlap,\\ Homogeneous effect}\\ \(\boldsymbol{\psi}_0=0.584\)}}
& GCOMP-GLM      & 0.567 & 0.017 & -2.96 & 0.020 & 0.011 & 0.018 & 0.018 \\
& GCOMP-SL       & 0.579 & 0.005 & -0.87 & 0.010 & 0.009 & 0.017 & 0.017 \\
& IPW-GLM        & 0.567 & 0.017 & -2.99 & 0.020 & 0.011 & 0.017 & 0.017 \\
& IPW-SL         & 0.575 & 0.010 & -1.63 & 0.014 & 0.010 & 0.011 & 0.011 \\
& IPW-GLM-OptPS  & 0.567 & 0.017 & -2.96 & 0.020 & 0.011 & 0.017 & 0.017 \\
& IPW-SL-OptPS   & 0.573 & 0.011 & -1.97 & 0.015 & 0.010 & 0.012 & 0.012 \\
& AIPW-GLM       & 0.567 & 0.017 & -2.98 & 0.020 & 0.011 & 0.017 & 0.017 \\
& AIPW-SL        & 0.581 & 0.003 & -0.59 & 0.010 & 0.009 & 0.006 & 0.006 \\
& TMLE-GLM       & 0.567 & 0.017 & -2.98 & 0.020 & 0.011 & 0.017 & 0.017 \\
& TMLE-SL        & 0.581 & 0.003 & -0.52 & 0.009 & 0.009 & 0.005 & 0.005 \\
\midrule

\multirow{10}{*}{\parbox{4.2cm}{\centering \textbf{Large overlap,\\ Heterogeneous effect}\\ \(\boldsymbol{\psi}_0=0.560\)}}
& GCOMP-GLM      & 0.542 & 0.019 & -3.36 & 0.021 & 0.010 & 0.021 & 0.021 \\
& GCOMP-SL       & 0.554 & 0.007 & -1.21 & 0.011 & 0.009 & 0.019 & 0.019 \\
& IPW-GLM        & 0.542 & 0.019 & -3.33 & 0.021 & 0.010 & 0.018 & 0.018 \\
& IPW-SL         & 0.550 & 0.010 & -1.82 & 0.014 & 0.010 & 0.011 & 0.011 \\
& IPW-GLM-OptPS  & 0.542 & 0.019 & -3.30 & 0.021 & 0.010 & 0.018 & 0.018 \\
& IPW-SL-OptPS   & 0.548 & 0.012 & -2.21 & 0.016 & 0.009 & 0.013 & 0.013 \\
& AIPW-GLM       & 0.542 & 0.019 & -3.32 & 0.021 & 0.010 & 0.018 & 0.018 \\
& AIPW-SL        & 0.557 & 0.004 & -0.68 & 0.009 & 0.008 & 0.006 & 0.006 \\
& TMLE-GLM       & 0.542 & 0.019 & -3.32 & 0.021 & 0.010 & 0.018 & 0.018 \\
& TMLE-SL        & 0.557 & 0.003 & -0.58 & 0.009 & 0.008 & 0.004 & 0.004 \\
\midrule

\multirow{10}{*}{\parbox{4.2cm}{\centering \textbf{Small overlap,\\ Homogeneous effect}\\ \(\boldsymbol{\psi}_0=0.584\)}}
& GCOMP-GLM      & 0.583 & 0.001 & -0.20 & 0.012 & 0.012 & 0.026 & 0.026 \\
& GCOMP-SL       & 0.581 & 0.003 & -0.47 & 0.011 & 0.010 & 0.027 & 0.027 \\
& IPW-GLM        & 0.589 & 0.004 & 0.75 & 0.017 & 0.016 & 0.030 & 0.030 \\
& IPW-SL         & 0.563 & 0.021 & -3.56 & 0.029 & 0.020 & 0.033 & 0.033 \\
& IPW-GLM-OptPS  & 0.587 & 0.003 & 0.52 & 0.015 & 0.014 & 0.027 & 0.027 \\
& IPW-SL-OptPS   & 0.564 & 0.020 & -3.44 & 0.028 & 0.020 & 0.033 & 0.033 \\
& AIPW-GLM       & 0.576 & 0.008 & -1.44 & 0.018 & 0.016 & 0.013 & 0.013 \\
& AIPW-SL        & 0.581 & 0.003 & -0.56 & 0.011 & 0.011 & 0.013 & 0.013 \\
& TMLE-GLM       & 0.581 & 0.003 & -0.55 & 0.014 & 0.013 & 0.009 & 0.009 \\
& TMLE-SL        & 0.584 & \biaszero & -0.11 & 0.011 & 0.011 & 0.002 & 0.002 \\
\midrule

\multirow{10}{*}{\parbox{4.2cm}{\centering \textbf{Small overlap,\\ Heterogeneous effect}\\ \(\boldsymbol{\psi}_0=0.560\)}}
& GCOMP-GLM      & 0.558 & 0.003 & -0.47 & 0.011 & 0.011 & 0.028 & 0.028 \\
& GCOMP-SL       & 0.556 & 0.004 & -0.78 & 0.011 & 0.010 & 0.028 & 0.028 \\
& IPW-GLM        & 0.561 & \biaszero & 0.14 & 0.016 & 0.016 & 0.035 & 0.035 \\
& IPW-SL         & 0.535 & 0.025 & -4.46 & 0.034 & 0.023 & 0.041 & 0.041 \\
& IPW-GLM-OptPS  & 0.560 & \biaszero & -0.10 & 0.014 & 0.014 & 0.034 & 0.034 \\
& IPW-SL-OptPS   & 0.536 & 0.024 & -4.28 & 0.032 & 0.021 & 0.041 & 0.041 \\
& AIPW-GLM       & 0.552 & 0.008 & -1.43 & 0.017 & 0.015 & 0.012 & 0.012 \\
& AIPW-SL        & 0.557 & 0.003 & -0.59 & 0.011 & 0.010 & 0.012 & 0.012 \\
& TMLE-GLM       & 0.558 & 0.003 & -0.47 & 0.013 & 0.013 & 0.007 & 0.007 \\
& TMLE-SL        & 0.560 & \biaszero & -0.10 & 0.010 & 0.010 & 0.002 & 0.002 \\
\bottomrule
\end{tabular}%
}
\end{table}

\begin{table}[H]
\centering
\caption{Point-estimation results for the W-oracle finite-sample sensitivity setting (\(n=500\)). The transformed covariates \(W=\phi(X)\), which drive the true treatment and outcome mechanisms, were supplied to the nuisance models.}
\label{tab:supp_w_n500}
\resizebox{\textwidth}{!}{%
\begin{tabular}{llrrrrrrr}
\toprule
Scenario & Estimator & Mean est. & Abs. bias & Bias (\%) & RMSE & Emp. SE & \(\theta_{\max}\) & \(p_{\max}\) \\
\midrule

\multirow{10}{*}{\parbox{4.2cm}{\centering \textbf{Large overlap,\\ Homogeneous effect}\\ \(\boldsymbol{\psi}_0=0.584\)}}
& GCOMP-GLM      & 0.583 & 0.001 & -0.23 & 0.018 & 0.018 & 0.002 & 0.003 \\
& GCOMP-SL       & 0.581 & 0.003 & -0.59 & 0.019 & 0.018 & 0.007 & 0.007 \\
& IPW-GLM        & 0.583 & \biaszero & -0.16 & 0.020 & 0.020 & 0.001 & 0.001 \\
& IPW-SL         & 0.576 & 0.008 & -1.39 & 0.023 & 0.021 & 0.008 & 0.007 \\
& IPW-GLM-OptPS  & 0.583 & 0.001 & -0.20 & 0.018 & 0.018 & 0.002 & 0.001 \\
& IPW-SL-OptPS   & 0.573 & 0.011 & -1.97 & 0.023 & 0.020 & 0.011 & 0.010 \\
& AIPW-GLM       & 0.583 & 0.001 & -0.20 & 0.018 & 0.018 & 0.002 & 0.002 \\
& AIPW-SL        & 0.582 & 0.003 & -0.43 & 0.018 & 0.018 & 0.004 & 0.004 \\
& TMLE-GLM       & 0.583 & 0.001 & -0.19 & 0.018 & 0.018 & 0.002 & 0.002 \\
& TMLE-SL        & 0.582 & 0.002 & -0.40 & 0.018 & 0.018 & 0.004 & 0.003 \\
\midrule

\multirow{10}{*}{\parbox{4.2cm}{\centering \textbf{Large overlap,\\ Heterogeneous effect}\\ \(\boldsymbol{\psi}_0=0.560\)}}
& GCOMP-GLM      & 0.559 & 0.001 & -0.20 & 0.017 & 0.017 & 0.002 & 0.003 \\
& GCOMP-SL       & 0.557 & 0.004 & -0.65 & 0.018 & 0.017 & 0.009 & 0.009 \\
& IPW-GLM        & 0.560 & \biaszero & -0.13 & 0.019 & 0.019 & 0.001 & 0.001 \\
& IPW-SL         & 0.552 & 0.008 & -1.49 & 0.022 & 0.020 & 0.008 & 0.008 \\
& IPW-GLM-OptPS  & 0.559 & \biaszero & -0.16 & 0.017 & 0.017 & 0.001 & 0.002 \\
& IPW-SL-OptPS   & 0.548 & 0.012 & -2.16 & 0.023 & 0.020 & 0.012 & 0.012 \\
& AIPW-GLM       & 0.559 & \biaszero & -0.16 & 0.017 & 0.017 & 0.002 & 0.002 \\
& AIPW-SL        & 0.558 & 0.003 & -0.47 & 0.017 & 0.017 & 0.004 & 0.004 \\
& TMLE-GLM       & 0.559 & \biaszero & -0.16 & 0.017 & 0.017 & 0.002 & 0.002 \\
& TMLE-SL        & 0.558 & 0.002 & -0.41 & 0.017 & 0.017 & 0.004 & 0.003 \\
\midrule

\multirow{10}{*}{\parbox{4.2cm}{\centering \textbf{Small overlap,\\ Homogeneous effect}\\ \(\boldsymbol{\psi}_0=0.584\)}}
& GCOMP-GLM      & 0.582 & 0.002 & -0.36 & 0.020 & 0.019 & 0.007 & 0.007 \\
& GCOMP-SL       & 0.576 & 0.008 & -1.36 & 0.022 & 0.020 & 0.018 & 0.018 \\
& IPW-GLM        & 0.579 & 0.005 & -0.88 & 0.034 & 0.033 & 0.008 & 0.008 \\
& IPW-SL         & 0.558 & 0.026 & -4.46 & 0.044 & 0.036 & 0.029 & 0.029 \\
& IPW-GLM-OptPS  & 0.579 & 0.005 & -0.89 & 0.030 & 0.029 & 0.011 & 0.011 \\
& IPW-SL-OptPS   & 0.560 & 0.024 & -4.18 & 0.040 & 0.032 & 0.029 & 0.029 \\
& AIPW-GLM       & 0.583 & 0.001 & -0.23 & 0.023 & 0.023 & 0.002 & 0.002 \\
& AIPW-SL        & 0.579 & 0.006 & -0.95 & 0.022 & 0.022 & 0.011 & 0.011 \\
& TMLE-GLM       & 0.578 & 0.006 & -1.03 & 0.026 & 0.025 & 0.009 & 0.009 \\
& TMLE-SL        & 0.578 & 0.006 & -1.05 & 0.024 & 0.023 & 0.011 & 0.011 \\
\midrule

\multirow{10}{*}{\parbox{4.2cm}{\centering \textbf{Small overlap,\\ Heterogeneous effect}\\ \(\boldsymbol{\psi}_0=0.560\)}}
& GCOMP-GLM      & 0.560 & \biaszero & -0.09 & 0.018 & 0.018 & 0.003 & 0.003 \\
& GCOMP-SL       & 0.553 & 0.007 & -1.25 & 0.021 & 0.019 & 0.016 & 0.016 \\
& IPW-GLM        & 0.555 & 0.005 & -0.98 & 0.035 & 0.034 & 0.009 & 0.009 \\
& IPW-SL         & 0.530 & 0.030 & -5.34 & 0.048 & 0.038 & 0.035 & 0.035 \\
& IPW-GLM-OptPS  & 0.555 & 0.005 & -0.98 & 0.031 & 0.030 & 0.011 & 0.011 \\
& IPW-SL-OptPS   & 0.532 & 0.028 & -5.03 & 0.044 & 0.034 & 0.035 & 0.035 \\
& AIPW-GLM       & 0.559 & \biaszero & -0.16 & 0.020 & 0.020 & 0.002 & 0.003 \\
& AIPW-SL        & 0.556 & 0.005 & -0.84 & 0.021 & 0.020 & 0.010 & 0.010 \\
& TMLE-GLM       & 0.555 & 0.006 & -1.04 & 0.024 & 0.023 & 0.010 & 0.010 \\
& TMLE-SL        & 0.555 & 0.005 & -0.98 & 0.023 & 0.022 & 0.011 & 0.011 \\
\bottomrule
\end{tabular}%
}
\end{table}

\begin{table}[H]
\centering
\caption{Point-estimation results for the X-working finite-sample sensitivity setting (\(n=500\)). The original baseline covariates \(X\), rather than the nonlinear transformed covariates \(W=\phi(X)\), were supplied to the nuisance models.}
\label{tab:supp_x_n500}
\resizebox{\textwidth}{!}{%
\begin{tabular}{llrrrrrrr}
\toprule
Scenario & Estimator & Mean est. & Abs. bias & Bias (\%) & RMSE & Emp. SE & \(\theta_{\max}\) & \(p_{\max}\) \\
\midrule

\multirow{10}{*}{\parbox{4.2cm}{\centering \textbf{Large overlap,\\ Homogeneous effect}\\ \(\boldsymbol{\psi}_0=0.584\)}}
& GCOMP-GLM      & 0.566 & 0.019 & -3.18 & 0.028 & 0.021 & 0.018 & 0.018 \\
& GCOMP-SL       & 0.576 & 0.008 & -1.41 & 0.021 & 0.019 & 0.020 & 0.020 \\
& IPW-GLM        & 0.566 & 0.018 & -3.13 & 0.028 & 0.021 & 0.015 & 0.015 \\
& IPW-SL         & 0.566 & 0.018 & -3.05 & 0.028 & 0.022 & 0.016 & 0.015 \\
& IPW-GLM-OptPS  & 0.566 & 0.018 & -3.16 & 0.028 & 0.021 & 0.015 & 0.015 \\
& IPW-SL-OptPS   & 0.563 & 0.021 & -3.65 & 0.030 & 0.021 & 0.019 & 0.018 \\
& AIPW-GLM       & 0.565 & 0.019 & -3.22 & 0.028 & 0.021 & 0.016 & 0.016 \\
& AIPW-SL        & 0.577 & 0.008 & -1.30 & 0.020 & 0.019 & 0.011 & 0.011 \\
& TMLE-GLM       & 0.565 & 0.019 & -3.22 & 0.028 & 0.021 & 0.016 & 0.016 \\
& TMLE-SL        & 0.577 & 0.007 & -1.22 & 0.020 & 0.019 & 0.008 & 0.008 \\
\midrule

\multirow{10}{*}{\parbox{4.2cm}{\centering \textbf{Large overlap,\\ Heterogeneous effect}\\ \(\boldsymbol{\psi}_0=0.560\)}}
& GCOMP-GLM      & 0.541 & 0.020 & -3.53 & 0.028 & 0.020 & 0.021 & 0.021 \\
& GCOMP-SL       & 0.551 & 0.009 & -1.65 & 0.020 & 0.018 & 0.022 & 0.022 \\
& IPW-GLM        & 0.541 & 0.019 & -3.42 & 0.028 & 0.020 & 0.018 & 0.018 \\
& IPW-SL         & 0.542 & 0.018 & -3.30 & 0.028 & 0.021 & 0.017 & 0.017 \\
& IPW-GLM-OptPS  & 0.541 & 0.019 & -3.46 & 0.028 & 0.020 & 0.018 & 0.018 \\
& IPW-SL-OptPS   & 0.538 & 0.022 & -3.95 & 0.030 & 0.021 & 0.020 & 0.020 \\
& AIPW-GLM       & 0.541 & 0.020 & -3.51 & 0.028 & 0.020 & 0.018 & 0.018 \\
& AIPW-SL        & 0.553 & 0.008 & -1.36 & 0.019 & 0.018 & 0.011 & 0.011 \\
& TMLE-GLM       & 0.541 & 0.020 & -3.51 & 0.028 & 0.020 & 0.018 & 0.018 \\
& TMLE-SL        & 0.553 & 0.007 & -1.23 & 0.019 & 0.017 & 0.008 & 0.008 \\
\midrule

\multirow{10}{*}{\parbox{4.2cm}{\centering \textbf{Small overlap,\\ Homogeneous effect}\\ \(\boldsymbol{\psi}_0=0.584\)}}
& GCOMP-GLM      & 0.582 & 0.002 & -0.37 & 0.023 & 0.023 & 0.027 & 0.027 \\
& GCOMP-SL       & 0.579 & 0.005 & -0.94 & 0.022 & 0.021 & 0.032 & 0.032 \\
& IPW-GLM        & 0.585 & 0.001 & 0.21 & 0.034 & 0.034 & 0.027 & 0.027 \\
& IPW-SL         & 0.555 & 0.029 & -4.93 & 0.045 & 0.035 & 0.039 & 0.039 \\
& IPW-GLM-OptPS  & 0.585 & \biaszero & 0.08 & 0.031 & 0.031 & 0.028 & 0.028 \\
& IPW-SL-OptPS   & 0.558 & 0.026 & -4.48 & 0.043 & 0.034 & 0.040 & 0.040 \\
& AIPW-GLM       & 0.573 & 0.012 & -1.99 & 0.054 & 0.052 & 0.014 & 0.014 \\
& AIPW-SL        & 0.578 & 0.006 & -0.98 & 0.023 & 0.023 & 0.019 & 0.019 \\
& TMLE-GLM       & 0.580 & 0.004 & -0.76 & 0.027 & 0.027 & 0.010 & 0.010 \\
& TMLE-SL        & 0.581 & 0.003 & -0.48 & 0.023 & 0.023 & 0.007 & 0.007 \\
\midrule

\multirow{10}{*}{\parbox{4.2cm}{\centering \textbf{Small overlap,\\ Heterogeneous effect}\\ \(\boldsymbol{\psi}_0=0.560\)}}
& GCOMP-GLM      & 0.557 & 0.003 & -0.57 & 0.023 & 0.023 & 0.029 & 0.029 \\
& GCOMP-SL       & 0.554 & 0.006 & -1.10 & 0.022 & 0.021 & 0.032 & 0.032 \\
& IPW-GLM        & 0.558 & 0.002 & -0.35 & 0.034 & 0.034 & 0.034 & 0.034 \\
& IPW-SL         & 0.527 & 0.034 & -6.04 & 0.050 & 0.037 & 0.049 & 0.049 \\
& IPW-GLM-OptPS  & 0.558 & 0.003 & -0.48 & 0.030 & 0.030 & 0.035 & 0.035 \\
& IPW-SL-OptPS   & 0.529 & 0.032 & -5.64 & 0.047 & 0.034 & 0.049 & 0.049 \\
& AIPW-GLM       & 0.550 & 0.011 & -1.90 & 0.052 & 0.051 & 0.013 & 0.013 \\
& AIPW-SL        & 0.555 & 0.005 & -0.92 & 0.023 & 0.022 & 0.018 & 0.018 \\
& TMLE-GLM       & 0.557 & 0.003 & -0.60 & 0.026 & 0.026 & 0.009 & 0.009 \\
& TMLE-SL        & 0.559 & 0.002 & -0.30 & 0.022 & 0.021 & 0.005 & 0.005 \\
\bottomrule
\end{tabular}%
}
\end{table}

\label{LastPage}
\end{document}